%% file: main.tex
\documentclass[conference]{IEEEtran}
\IEEEoverridecommandlockouts

\usepackage{cite}
\usepackage{amsmath,amssymb,amsfonts}
\usepackage{algorithmic}
\usepackage{graphicx}
\usepackage{textcomp}
\usepackage{xcolor}
\def\BibTeX{{\rm B\kern-.05em{\sc i\kern-.025em b}\kern-.08em
    T\kern-.1667em\lower.7ex\hbox{E}\kern-.125emX}}
\usepackage[many]{tcolorbox}
\tcbuselibrary{listings,breakable}
\usepackage{xcolor}
\usepackage[utf8]{inputenc}
\DeclareUnicodeCharacter{2248}{$\approx$}
\usepackage{lipsum}
\usepackage[caption=false,font=footnotesize]{subfig}
\usepackage{booktabs}
\usepackage{multirow}
\usepackage{siunitx}
\usepackage{tabularx}
\usepackage{placeins}
\usepackage{listings}
\usepackage{xcolor}
\usepackage{caption}
\usepackage{float}

\floatname{listing}{Listing}
\newfloat{listing}{t}{lop}
\floatname{listing}{Listing}
\definecolor{codebg}{rgb}{0.95,0.95,0.95}
\definecolor{codeborder}{rgb}{0.7,0.7,0.7}
\definecolor{codecomment}{rgb}{0.0,0.5,0.0}
\definecolor{codekeyword}{rgb}{0.0,0.0,0.7}
\definecolor{codestring}{rgb}{0.6,0.0,0.0}

\lstdefinestyle{mystyle}{
  backgroundcolor=\color{codebg},
  frame=single,
  rulecolor=\color{codeborder},
  framerule=0.4pt,
  basicstyle=\ttfamily\tiny,
  keywordstyle=\color{codekeyword}\bfseries,
  commentstyle=\color{codecomment}\itshape,
  stringstyle=\color{codestring},
  breaklines=true,
  columns=fullflexible,
  keepspaces=true,
  showstringspaces=false,
  aboveskip=0pt,
  belowskip=0pt,
  lineskip=-1pt,
  framesep=2pt,
  xleftmargin=2pt,
  xrightmargin=2pt,
  framexleftmargin=2pt,
  framexrightmargin=2pt
}
\lstdefinestyle{promptstyle}{
  basicstyle=\ttfamily\tiny,
  breaklines=true,
  columns=fullflexible,
  keepspaces=true,
  showstringspaces=false,
  aboveskip=0pt,
  belowskip=0pt,
  lineskip=-1pt
}
\usepackage[T1]{fontenc}
\usepackage[utf8]{inputenc}
\usepackage{tikz}
\usepackage{hyperref} 
\usetikzlibrary{patterns}
\definecolor{cachegray}{HTML}{9EA4AA}
\definecolor{cachegold}{HTML}{B8861B}
\definecolor{baselinegray}{HTML}{555555}
\newcommand{\legbox}[1]{%
\tikz{
\fill[cachegold!75] (0,0) rectangle (0.24,0.15);
\draw[pattern=#1,pattern color=black] (0,0) rectangle (0.24,0.15);
\draw[line width=0.35pt] (0,0) rectangle (0.24,0.15);
}
}

\newcommand{\legboxgray}{%
\tikz\draw[fill=cachegray!85,draw=black,line width=0.35pt]
(0,0) rectangle (0.24,0.15);
}
\usepackage[ruled,vlined]{algorithm2e}
\usepackage{xspace}
\def\repjudge{KD-Judge\xspace}
\usepackage{ulem}
\definecolor{myred}{RGB}{0,0,0}
\definecolor{cams}{RGB}{223,3,25}
\definecolor{cus}{RGB}{223,3,252}
\definecolor{cus1}{RGB}{0,0,0}
\newcommand{\mycut}[1]{}
\newcommand{\change}[1]{\textcolor{myred}{#1}}

\newcommand{\fan}[1]{\textcolor[rgb]{0,0,0}{#1}}
\newcommand{\shaibal}[1]{\textcolor{cus1}{#1}}
\begin{document}

\title{KD-Judge: A Knowledge-Driven Automated Judge Framework for Functional Fitness Movements on Edge Devices\\
}
\author{
\IEEEauthorblockN{
Shaibal Saha\IEEEauthorrefmark{1}\IEEEauthorrefmark{3},
Fan Li\IEEEauthorrefmark{1}\IEEEauthorrefmark{3},
Yunge Li\IEEEauthorrefmark{1},
Arun Iyengar\IEEEauthorrefmark{2},
Lucas Alves\IEEEauthorrefmark{1},
Lanyu Xu\IEEEauthorrefmark{1}
}

\IEEEauthorblockA{\IEEEauthorrefmark{1}
Department of Computer Science and Engineering, Oakland University, Rochester Hills, USA
}
\IEEEauthorblockA{\IEEEauthorrefmark{2}
Intelligent Data Management and Analytics LLC, USA
}


}

\maketitle
\begingroup
\renewcommand\thefootnote{\ddag}
\footnotetext{\parbox[t]{\linewidth}{Indicates equal contribution; order determined by a coin toss.\\
Corresponding authors: Shaibal Saha and Fan Li.\\
E-mail: \{shaibalsaha,fanli\}@oakland.edu.
}}
\endgroup
\begin{abstract}
Functional fitness movements are widely used in training, competition, and health-oriented exercise programs, yet consistently enforcing repetition (rep) standards remains challenging due to subjective human judgment, time constraints, and evolving rules. Existing AI-based approaches mainly rely on learned scoring or reference-based comparisons and lack explicit rule-based approaches, limiting transparency and deterministic rep-level validation. To address these limitations, we propose \textbf{\repjudge}, a novel knowledge-driven automated judging framework for functional fitness movements. It converts unstructured rulebook standards into executable, machine-readable representations using an LLM-based retrieval-augmented generation and chain-of-thought rule-structuring pipeline. The structured rules are then incorporated by a deterministic rule-based judging system with pose-guided kinematic reasoning to assess rep validity and temporal boundaries. To improve efficiency on edge devices, including a high-performance desktop and the resource-constrained Jetson AGX Xavier, we introduce a dual strategy caching mechanism that can be selectively applied to reduce redundant and unnecessary computation. Experiments demonstrate reliable rule-structuring performance and accurate rep-level assessment, with judgment evaluation conducted on the CFRep dataset, achieving faster-than-real-time execution (real-time factor (RTF) $<$ 1). When the proposed caching strategy is enabled, the system achieves up to 3.36$\times$ and 15.91$\times$ speedups on resource-constrained edge device compared to the non-caching baseline for pre-recorded and live-streaming scenarios, respectively. These results show that \repjudge enables transparent, efficient, and scalable rule-grounded rep-level analysis that can complement human judging in practice.
\end{abstract}

\input{Sections/Introduction}

\input{Sections/related_work}

\input{Sections/methodology}
\input{Sections/results}
\input{Sections/conclusion}
\section*{Acknowledgment}
This work was supported in part by the NSF under award \#2245729.

\bibliographystyle{IEEEtran}
\bibliography{reference}

\end{document}

%% file: Sections/Introduction.tex
\section{Introduction}
In recent years, functional fitness sports have become popular worldwide by integrating strength, endurance, and conditioning into highly efficient training structures that enhance physical fitness. \fan{These advantages have attracted more people to participate in training and have led to the development of large-scale competitions.}
Functional fitness competitions consist of multiple movements performed in sequence. Each is defined by strict and movement-specific repetition (rep) completion standards. These standards are fundamental to the competition, as they directly determine the validity of each rep under strict temporal and kinematic boundaries.

\change{As a result, the core challenge in functional fitness competitions lies not only in performing the movements, but in consistently enforcing rep completion standards during judging. This requirement reframes functional fitness judging as the execution of explicit, rulebook-defined constraints over high-frequency human motion data, rather than a subjective scoring task.} Traditionally, these rep standards are enforced by human judges; however, judging in high-stakes functional fitness competitions imposes cognitive demands. \shaibal{Human judges must assess complex movements over very short time intervals, where perceptual and cognitive constraints can introduce variability in rep validity decisions.} \change{Moreover, these challenges are compounded by the fact that functional fitness standards evolve over time across competitions, requiring judging systems to adapt to rule changes in a systematic manner.}

Beyond competitive settings, the same rep-completion standards are important in health-oriented exercise monitoring, where incorrect reps can increase injury risk and undermine training effectiveness. Consequently, automated rule-based rep validity assessment directly supports connected health applications, including exercise monitoring, injury prevention, and personalized training.

\change{Meanwhile, recent advances in artificial intelligence have enabled automated analysis of human motion across a wide range of applications ~\cite{van2021machine, halilaj2018machine, harris2022survey, sepas2022deep, morshed2023human}. In competitive sports contexts, prior work has explored fine-grained motion analysis and performance scoring~\cite{xu2022finediving, shao2020finegym}. However, these approaches primarily rely on learned scoring or recognition models and do not support deterministic rep validity assessment. In contrast, other studies focus on reconstructing 3D human pose from video and comparing athlete executions against professional instructor references to provide feedback~\cite{fieraru2021aifit}.} Despite their methodological differences, automated sports judging systems do not explicitly encode competition rules, are difficult to adapt to evolving standards, and fail to provide transparent, auditable rep-level decisions required for functional fitness competitions~\cite{lewis2020retrieval}. 
\begin{figure}[!t]
  \centering
  \subfloat[]{
    \includegraphics[width=0.468\linewidth]{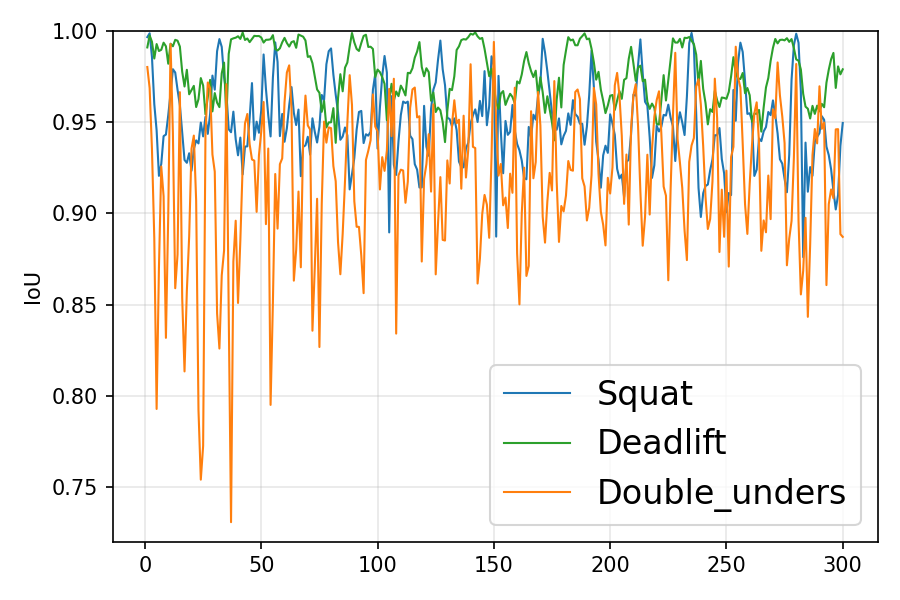}
    \label{fig:det_iou}
  }
  \hfill
  \subfloat[]{
    \includegraphics[width=0.462\linewidth]{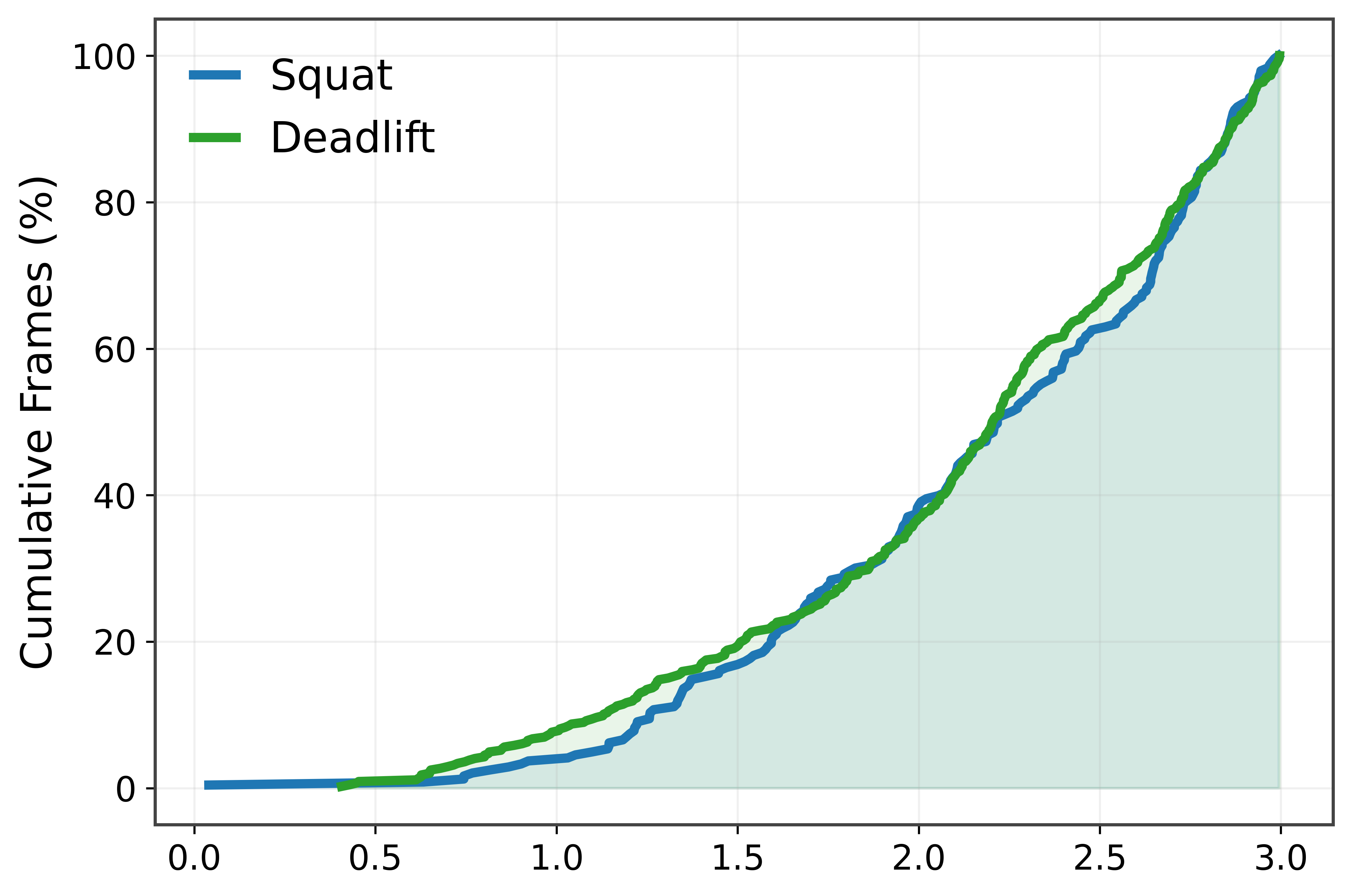}
     \label{fig:roi_mot}
  }
\caption{Redundancy analysis motivating dual strategy cache mechanism. (a) Detection bbox IoU across frames (x: frame index). (b) CDF of frame-wise ROI pixel-wise difference between consecutive frames (x: difference value).}
\label{fig:cache_motiv}
\end{figure}
Given the current state of functional fitness practice and competition, there is a growing demand for an automated judging system. Such a system should be capable of interpreting competition rules, applying movement-specific standards, and evaluating rep validity and temporal boundaries. \fan{However, reliable rep-level judging with pose-based analysis (using top-down or bottom-up pose models) introduces significant computational overhead. In practice, video streams are processed frame by frame, and consecutive frames often exhibit strong temporal similarity, which leads to redundant computation. This computational burden becomes a key bottleneck when the system is deployed in resource-constrained devices. As shown in Figure~\ref{fig:det_iou}, the bounding box (bbox) intersection over Union (IoU) between adjacent frames in a top-down pose model is typically high, indicating that repeated detector invocation can be avoided. Figure~\ref{fig:roi_mot} shows that the cumulative distribution function (CDF) of frame-wise region-of-interest (ROI) appearance differences is concentrated at low values, indicating that many consecutive frames exhibit only minor visual changes and suggesting substantial temporal redundancy across frames.}

Therefore, designing a knowledge-driven automated judging framework capable of efficient deployment on edge devices requires addressing several challenges.
\textbf{Challenge 1}: How to extract relevant information from rulebooks and translate it into machine-interpretable representations. \textbf{Challenge 2}: How to utilize the extracted knowledge to guide movement evaluation. \textbf{Challenge 3}: How to efficiently process video data by reducing redundant and unnecessary computation, enabling more effective deployment on edge devices, particularly in resource-constrained devices.

To address these challenges, we propose \textbf{\repjudge}, a novel knowledge-driven automated judging framework for functional fitness. The framework operates on multimodal inputs, including rulebook text and video, and follows a three-stage design. In the first two stages, an LLM-assisted custom-designed retrieval-augmented generation (RAG)~\cite{lewis2020retrieval}, and chain-of-thought (CoT) reasoning~\cite{wei2022chain} pipeline converts unstructured standards into machine-readable structured representations. In the final stage, these rules are executed by a deterministic rule-based judging system that uses pose-derived kinematic features to determine rep validity and temporal boundaries. Because rules are derived from the rulebook text rather than being hard-coded, \repjudge remains adaptable to changes in competition standards without altering the core judging architecture. Meanwhile, to reduce efficiency bottlenecks on resource-constrained edge devices, we introduce two caching strategies: detector caching (DC) and ROI temporal caching (RTC). DC skips redundant detector runs by reusing the cached bbox. RTC skips pose inference for visually similar frames by reusing cached pose results. To the best of our knowledge, this is the \textbf{first rule-based, knowledge-driven judging system} for functional fitness movements. It is designed for efficient deployment on edge devices, including a high-performance desktop and the resource-constrained Jetson AGX Xavier, and produces transparent, rule-based decisions rather than learned score predictions.

Our experiments demonstrate that the RAG-based rule-structuring pipeline correctly extracts and formats movement rules, with outputs aligned with human-in-the-loop experts and an independent LLM. Building on the structured rules, we evaluate the \repjudge judging system on the CFRep dataset~\cite{alves2025repval} and demonstrate accurate and efficient rep validity assessment across multiple movements and camera views. The system achieves faster-than-playback evaluation (real-time factor (RTF)$<$ 1) while maintaining reliable movement accuracy. We further analyze the efficiency of \repjudge, where the proposed caching strategies provide up to $3.36\times$ and $15.91\times$ speedups on resource-constrained edge devices over the system without caching for prerecorded and live-streaming scenarios, respectively, while preserving rep-level judging accuracy. Together, these results demonstrate that \repjudge supports efficient and scalable rep-level analysis under practical computational constraints. Our main contributions can be summarized as follows:
\begin{itemize}
\item We propose \textbf{\repjudge}, the first knowledge-driven automated judging framework for functional fitness that combines an LLM-assisted rule-structuring pipeline with a deterministic rule-based judging system to perform transparent, rep validity assessment from multimodal inputs.
\item To ensure the efficiency of \repjudge on edge devices,  particularly in resource-constrained devices, we design a \textbf{dual strategy caching mechanism} (DC and RTC) in the judging system that can be selectively enabled to reduce redundant computation and support efficient deployment.
\item We conduct a comprehensive evaluation that validates the quality of rule extraction and evaluates rep-level accuracy and efficiency of the judging system on the CFRep dataset under both prerecorded and live-streaming settings on edge devices.
\end{itemize}

%% file: Sections/related_work.tex
\section{Related Work}
\noindent\textbf{LLMs and RAG for Structured Rule Extraction.}
Recent research has explored LLM-based systems that integrate RAG to transform unstructured data into actionable representations. In sports analytics, this paradigm has been applied to multimodal narrative generation in soccer~\cite{10943875} and tactical analysis in badminton~\cite{BHARADWAJ2026115027}, where vision-based events are structured into queryable knowledge bases. For motion analysis, SportsGPT~\cite{tian2025sportsgpt} utilized RAG to align skeletal motion sequences with reference models for interpretable training guidance. Similarly, RAG-HAR~\cite{sivaroopan2025rag} demonstrates training-free activity recognition via retrieval over lightweight descriptors. 

\change{Recent work has also explored extracting procedural knowledge from unstructured documents for downstream verification and reasoning~\cite{polak2024extracting,elnashar2025enhancing,prompv}. CheckManual~\cite{long2025checkmanual}, for example, extracts procedural rules from manuals, demonstrating the feasibility of LLM-assisted rule extraction from textual sources. However, such approaches remain confined to textual or symbolic representations and do not ground extracted rules in sensor-derived signals required for deterministic, kinematics-based validation. We address this gap by integrating RAG and CoT reasoning to extract movement standards from rulebooks and map them to a pose-based keypoint schema, enabling rep-level validity assessment from video. }
\mycut{Complementary work proposed structured output prompting and document-to-structure pipelines, showing that prompt style and schema constraints affect extraction accuracy and cost, and enabling reliable conversion of unstructured representations into structured representations~\cite{polak2024extracting,elnashar2025enhancing,prompv}. One of the closest works like ours, named CheckManual~\cite{long2025checkmanual}, extracts procedural rules from manuals for downstream verification. However, these methods largely remain in textual or symbolic domains and do not ground extracted rules in sensor-derived signals required for deterministic, kinematics-based validation. We address this gap by integrating RAG and CoT reasoning to extract movement standards from rulebooks and map them to a pose-based keypoint schema, enabling deterministic execution for rep validity assessment from raw video.}
\begin{figure}[]
  \centering
  \includegraphics[width=\linewidth]{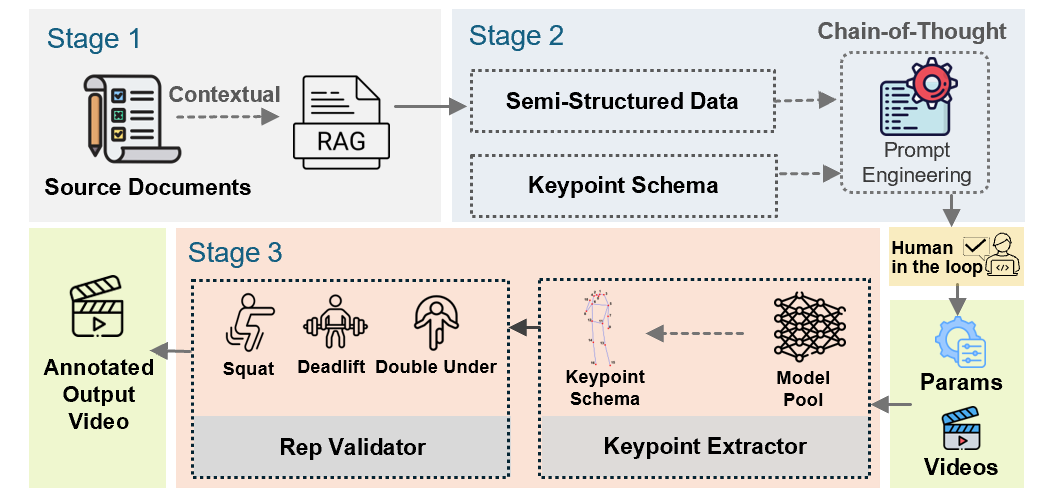}
    \caption{Overview of the proposed \repjudge framework. (Top) Stages~1–2 generate structured rules from unstructured standards. (Bottom) Stage~3 employs a rule-based judging system to determine rep validity and rep boundaries.}
  \label{fig:overview}
\end{figure}

\noindent\textbf{Action Quality Assessment.}
Action Quality Assessment (AQA) evaluates the execution quality of actions and is widely applied in sports analysis for fine-grained performance evaluation.
FineDiving~\cite{xu2022finediving} employs hierarchical temporal modeling with segmentation-based attention to emphasize informative subaction segments for action quality assessment.
To address uncertainty in figure skating assessment, LUSD-Net~\cite{xu2022finediving} disentangles technical and presentation-level action representations.
AGF-Olympics~\cite{zahan2024learning} introduces a non-local attention–based model that transforms dense features into sparse representations for score regression in artistic gymnastics floor routines.
Alternatively, from a multimodal perspective, LucidAction~\cite{dong2024lucidaction} provides multi-view RGB videos along with 2D and 3D pose sequences and adapts several unimodal AQA architectures for comprehensive sports action assessment. 
Unlike prior AQA methods for Olympic sports or gymnastics that focus on score prediction, our work targets functional fitness and evaluates rep validity using automated rule structuring and a deterministic judging system.

\noindent\textbf{Pose-based Judging.}
\change{Prior work has explored pose-based analysis for sports judging and fitness assessment using both learned and rule-based approaches. AI Coach \cite{wang2019ai} applies a pose estimation network to extract 2D human keypoints and uses kinematic features to train a supervised classifier for detecting predefined bad poses in freestyle skiing aerials. Subsequently, AIFit \cite{fieraru2021aifit} performs fitness assessment by reconstructing 3D human pose, and comparing trainee executions with professional instructor references to provide feedback. These methods rely on either learned classifiers or reference demonstrations, which limit interpretability and adaptability when movement standards change. In contrast, our approach is knowledge-driven, using LLMs to extract structured rules from rulebooks and evaluating pose-derived kinematic features with deterministic rule-based functions, enabling transparent rep validation without instructor demonstrations or learned classifiers. 
\begin{figure}[]
  \centering
  \includegraphics[width=\linewidth]{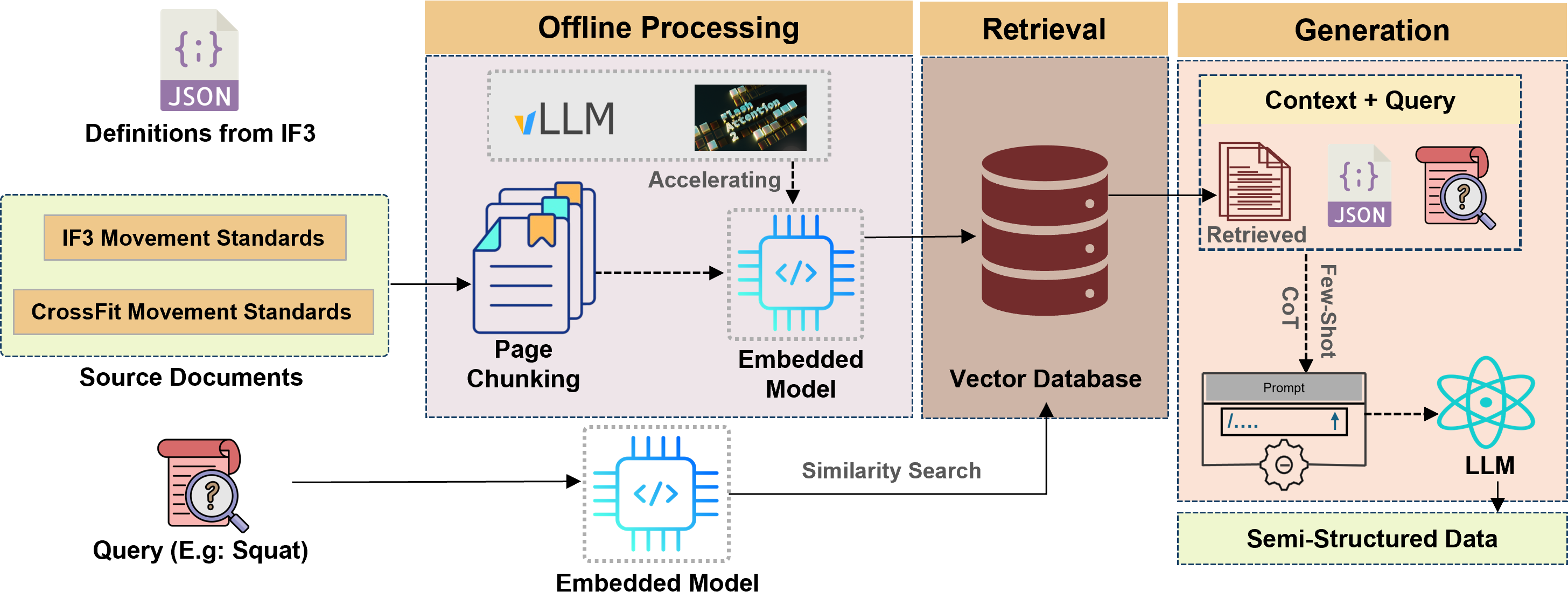}
  \caption{Overview of Stage~1 RAG pipeline for converting unstructured documents into semi-structured representations.}
  \label{fig:stage1}
\end{figure}
More closely related to our setting, RepVal \cite{alves2025repval} uses MediaPipe to extract 2D human poses and applies rule-based functions to manually designed kinematic features for rep validation, making it less flexible than our knowledge-driven approach. Additionally, RepVal uses a unified threshold across models and camera views and evaluates performance at the sequence level, without accounting for rep-level temporal alignment. However, \repjudge addresses these issues by adopting adaptive thresholds and using rep boundaries to avoid temporal misalignment in metric evaluation.}

%% file: Sections/methodology.tex
\section{\repjudge}
Figure~\ref{fig:overview} illustrates an overview of the proposed \repjudge framework, which converts unstructured movement standards into structured representations for automated judgment of functional fitness movements. The framework follows a three-stage design. In Stages~1 and~2, unstructured movement standards are processed using a RAG pipeline~\cite{lewis2020retrieval} to obtain semi-structured representations, which are then aligned with a predefined keypoint schema using CoT reasoning to produce structured movement representations. In Stage~3, input videos and the structured movement representation, along with a set of parameters are used to evaluate the rep validity. 

\subsection{Stage~1: Unstructured to Semi-Structured}
To address Challenge 1 and support evolving rulebooks without manual rule encoding, \repjudge designs a resource-efficient RAG pipeline to transform unstructured movement standards into semi-structured representations, incorporating quantized embeddings and source-aware retrieval thresholds for heterogeneous source documents. As illustrated in Figure~\ref{fig:stage1}, relevant source documents are first retrieved using similarity search. The retrieved texts and additional movement definitions are used as context in the generation phase. 
\subsubsection{Document Ingestion and Preprocessing} 
We utilize two primary source documents for Stage~1: the 2025 IF3 movement standards published by the International Functional Fitness Federation (IF3)\footnote{\href{https://functionalfitness.sport/wp-content/uploads/2025/01/iF3-Movement-standards-1.pdf}{IF3 Movement Standards (2025)}}
 and the CrossFit competition rulebook\footnote{\url{https://games.crossfit.com/rules}}. As these documents differ substantially in structure and content organization, we apply document-specific preprocessing to normalize formatting and remove non-movement sections. Each page is assigned a persistent source label (IF3: \texttt{label=1}, CrossFit: \texttt{label=0}) that is retained throughout chunking, embedding, and retrieval to enable source-aware filtering. In addition, key movement definitions are extracted separately and injected during generation to ensure consistent interpretation of movement terminology.
\begin{listing}[]
\tiny
\vspace{-0.8em}
\caption{Prompt template for Stage~1.}
\label{lst:movement_prompt}
\centering
\begin{minipage}{0.98\linewidth}
\begin{lstlisting}[style=mystyle, language=Python]
# You are an expert judge for functional fitness movements.
# Task: extract movement standards and output STRICT JSON.
{
  "movement": "",
  "response": {
    "rep_start": ["..."],
    "rep_end": ["..."],
    "rep_requirements": ["..."],
    "no_rep_conditions": ["..."]
  },
}
# Context:
# -- Retrieved pages from IF3 and CrossFit Rulebook
# -- Definition for shorthand terms (e.g., "Standing Tall", "Bottom-of-Squat")
# Instructions:
# 1. Use EXACT wording from the retrieved movement text.
# 2. Expand shorthand terms using their full definitions.
# 3. Do NOT hallucinate or infer conditions not stated.
# 4. Maintain JSON validity -- no extra commentary.

# Example (Few-Shot Guidance)
Input: "The athlete must start with knees fully extended..."
Output:
{
  "movement": "Air Squat",
  "response": {
    "rep_start": {"bodypart": "condition",...},
    "rep_end": {"bodypart": "condition",...},
    "rep_requirements": {"bodypart": "condition",...},
    "no_rep_conditions": ["Heels off the ground",..]
  },
}
\end{lstlisting}
\end{minipage}
\end{listing}
\subsubsection{Chunking Strategy and Embedding Model}
We adopt page-level chunking strategy in Stage~1, where each chunk $C_i$ corresponds to a single page. This page-level chunking aligns with the inherent structure of both source documents, where each page typically encapsulates a complete movement. Compared to sentence-level or overlapping chunking, this strategy provides stable retrieval boundaries and preserves the full context of each movement description. This makes it well-suited to the structure of the source documents. Each chunk is associated with its metadata fields (\texttt{label}, \texttt{sourceType}, and \texttt{pageIndex}), ensuring traceability across the embedding and retrieval phase. 

Each $C_i$ is converted into a dense vector representation $\mathbf{v}_i \in \mathbb{R}^d$, where $d$ denotes the embedding dimension. To balance retrieval performance and resource efficiency, \repjudge implements post-training quantization on Linq-Mistral embedding model~\cite{choi2024linq} using activation-aware Weight quantization (AWQ)~\cite{lin2024awq}. The resulting model is deployed via the vLLM inference backend~\cite{kwon2023efficient} with FlashAttention-2~\cite{dao2023flashattention} enabled to accelerate attention computation during embedding generation. The resulting embeddings are stored in vector database as key–value entries, $(\mathbf{v}_i, M_i, T_i) \in D_{\text{redis}}$, where $M_i$ contains the associated metadata fields, and $T_i$ stores the corresponding textual content of the chunk.  
\subsubsection{Vector Storage and Similarity-Based Retrieval}
Given an input query, the text is embedded in a dense vector representation $\mathbf{q} \in \mathbb{R}^d$ using the same quantized model to maintain embedding consistency. Retrieval is then performed by computing cosine similarity between the query embedding and indexed document embeddings using Eq.~\ref{cosine_similarity}, where $s_i$ measures the semantic similarity between the query and the embedding $\mathbf{v}_i$ of chunk $C_i$. Each stored embedding is retrieved along with its associated metadata, which guides source-specific filtering and thresholding during similarity search.
\begin{equation}
\label{cosine_similarity}
s_i = \frac{\mathbf{q} \cdot \mathbf{v}_i}{\|\mathbf{q}\|\,\|\mathbf{v}_i\|}, \quad s_i \in [0, 1]
\end{equation}
Because source documents follow different structural patterns, we use label-specific cosine similarity thresholds. These thresholds are chosen empirically (Section~\ref{threshold_analysis}) to balance retrieval precision and recall. For a given query, top-$k$ candidates are first retrieved based on cosine similarity and the source label. Then, threshold filtering is applied—$0.4$ for the IF3 rulebook and $0.6$ for the CrossFit rulebook to discard low-confidence matches. This selective filtering improves retrieval accuracy by ensuring that only the most contextually relevant pages from each source are retained for generation. The retrieved content then serves as context for the LLM to generate semi-structured representations.

\subsubsection{Prompt Engineering}
Once the relevant documents are retrieved, \repjudge applies a prompt-engineered generation strategy that combines the query, retrieved documents, structured IF3 definitions, few-shot examples, and CoT guidance (see Listing~\ref{lst:movement_prompt}) to generate semi-structured representations in the generation phase. The IF3 definitions are provided as additional context to resolve shorthand expressions in the rulebooks and ensure consistent interpretation across movements. We employ GPT-4.1~\cite{achiam2023gpt} as the generation model due to its strong reasoning capability.
\begin{listing}[]
\vspace{-0.68em}
\tiny
\caption{Prompt template for Stage~2.}
\label{lst:movement_prompt1}
\centering
\begin{minipage}{0.98\linewidth}
\begin{lstlisting}[style=mystyle, language=Python]
You are a biomechanics and AI movement standard expert.
Convert the following semi-structured rules for "{movement_name}" into structured JSON for a pose-evaluation engine.

## Known Keypoints (COCO WholeBody)
{", ".join(joint_list)}

## Conversion Rules:
- Vertical alignment -> X(joint1) ~= X(joint2)
- Depth / below-plane -> Y(jointA) < Y(jointB)
- Extension -> Angle(jointA, jointB, jointC) ~= 180 deg
- Partial bend -> Angle(jointA, jointB, jointC) < 180 deg
- Symmetry -> add both sides with 'and'
- Use semantic keys (e.g., vertical_alignment, hip_extension)

## Example:
Input: "hips and knees fully extended"
Output:
{
  "hip_knee_extension": {
    "keypoints": ["left_hip","right_hip","left_knee","right_knee"],
    "condition": "Angle(...) ~= 180 deg"
  }
}
\end{lstlisting}
\end{minipage}
\end{listing}
\subsection{Stage 2: Semi-structured to Structured}
Building on the semi-structured representations from Stage~1, \repjudge completes the translation objective discussed in Challenge~1 and establishes the kinematic grounding necessary to address Challenge~2. To construct machine-readable representations, \repjudge converts the semi-structured representations from Stage~1 into structured representations aligned with a pose-based keypoint schema, enabling direct interpretation by the judge system. The keypoint schema is treated as a coordinate reference system over human body joints. Qualitative movement descriptions are then decomposed into geometric constraints on these joints. Each movement is represented as a collection of spatial and kinematic constraints grounded in the keypoint schema~\cite{mmpose2020}.
\begin{figure}[]
  \centering
  \includegraphics[width=0.95\linewidth]{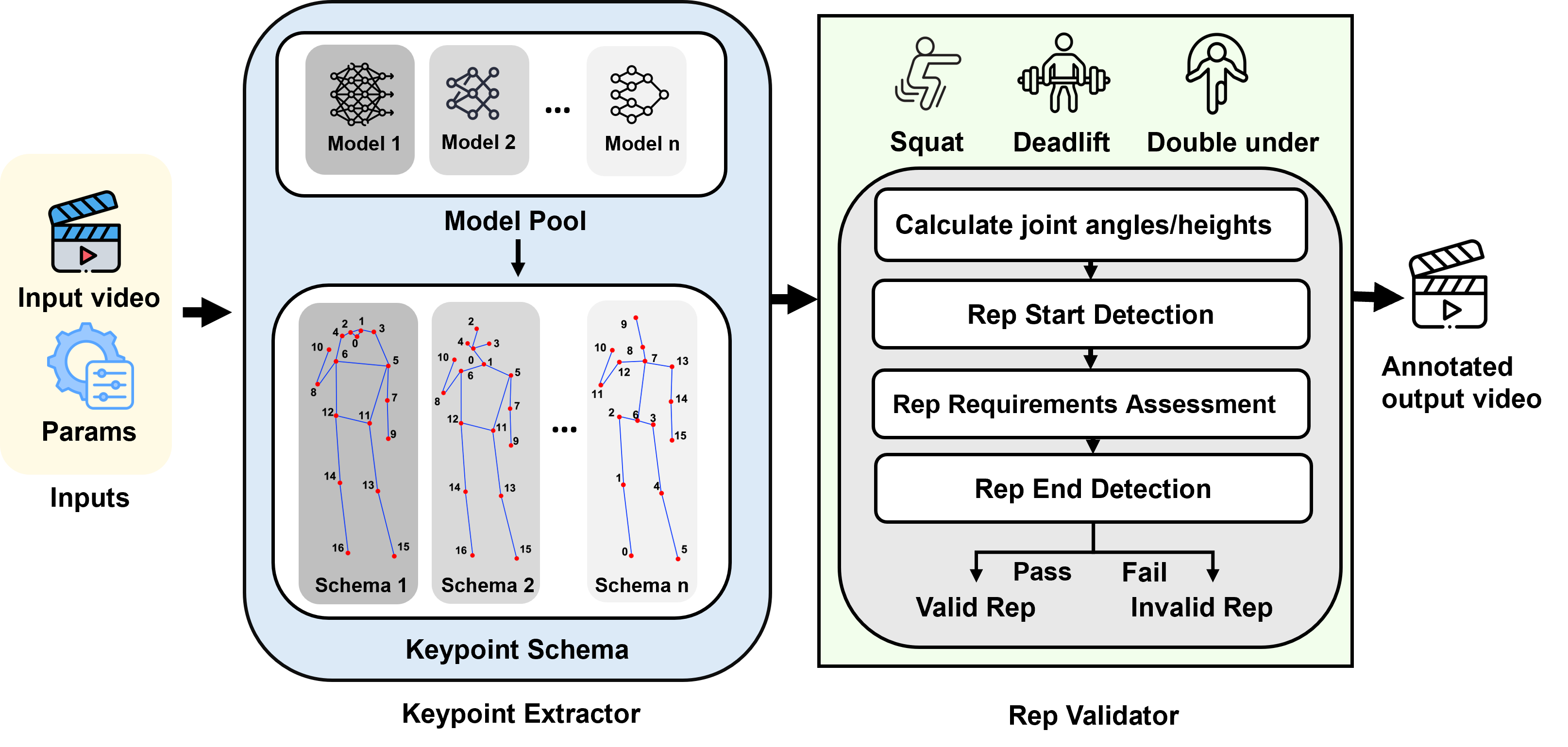}
    \caption{Overview of Stage 3 Judge System: The judge system extracts keypoints and evaluates rep validity and boundaries.} 
  \label{fig:judeg_system}
\end{figure}

The conversion employs an LLM-guided schema grounding strategy with CoT reasoning (see Listing~\ref{lst:movement_prompt1}). We use GPT-4.1 with low-temperature decoding (temperature = 0.1) to generate pose geometric and kinematic constraints from the semi-structured representations in a consistent manner. During this process, requirements are mapped to explicit geometric relations over the available keypoints. This includes $X$-axis alignment constraints for verticality, $Y$-axis inequalities for depth and ordering, and angular relations for joint extension (e.g., $\texttt{Angle}(\text{hip}, \text{knee}, \text{ankle}) \approx 180^\circ$ for full extension). Equality and inequality operators encode alignment and spatial ordering between joints, while trigonometric relations represent extension and flexion conditions. Before passing the structured representations to Stage~3, a human-in-the-loop identifies and verifies key rules from the structured representations. For instance, in the case of full-body extension, the structured representation derived from the standard rulebook includes hip, knee, and ankle angles. Once the hip and knee angles reach the full-extension range, the body is biomechanically upright, making the computation of the ankle angle redundant. A human-in-the-loop removes such unnecessary computations while preserving biomechanical correctness.

\subsection{Stage 3: Judge System}
Our judge system is designed to evaluate the validity of movement reps. The system consists of two main components: a keypoint extractor and a rep validator (Figure~\ref{fig:judeg_system}). The system inputs include a video sequence and a set of parameters, which contain predefined thresholds, structured representations generated by Stage 2, and system settings such as the selected pose model and movement type. \shaibal{To ensure real-time viability on edge devices, Stage~3 incorporates a selectively enabled dual strategy caching mechanism. This system targets computational redundancy through a DC and a universal RTC caching mechanism, enabling it to maintain high judging throughput.}


\subsubsection{Keypoint extractor}
In the keypoint extractor, we integrate MMPose~\cite{mmpose2020} and maintain a model pool that provides multiple 2D pose estimation models trained on different datasets (see Table~\ref{tab:pose_models}). Each selected model outputs keypoints according to its own schema. The selection of pose models can be controlled through the input parameters. The provided models can be categorized as top-down and bottom-up \cite{zhou2023human}. Top-down pose estimation detects people first and then estimates poses within each bounding box, whereas bottom-up methods detect all keypoints first and then group them into individual persons.
Since the model pool consists of 2D multi-person pose estimation models, and the videos contain multiple individuals, person tracking is required to consistently follow the target subject. At initialization, the target person can be identified by occupying a large area in the frame. However, during exercise movements, other individuals may temporarily occupy a larger portion of the frame. To handle such cases and avoid identity switches, we integrate tracking algorithms into the system. Once the target person’s track ID is determined, this ID is maintained and tracked across subsequent frames. \repjudge adopts two widely used tracking strategies: IoU-based and Object keypoint similarity (OKS)-based tracking.

\noindent \textbf{IoU-based Tracking.} 
Object identities are associated across frames by IoU-based bounding box matching, where instances inherit track IDs if the IoU exceeds a threshold and otherwise start new tracks. This strategy is used for top-down pose estimation models that provide person bounding boxes.

\noindent \textbf{OKS-based Tracking.} OKS adopts a distance-based formulation for keypoints. Keypoint similarity is computed in a keypoint-wise manner as shown in Eq.~\ref{oks}.
\vspace{-2mm}
\begin{equation}
\label{oks}
\mathrm{OKS}
=
\frac{
\sum_{i=1}^{K}
\exp\!\left(
-\frac{d_i^2}{2 s^2 \kappa_i^2}
\right)
\cdot \delta(v_i > 0)
}{
\sum_{i=1}^{K} \delta(v_i > 0)
}
\end{equation}
Each pose instance is represented by $K$ keypoints. $d_i$ denotes the Euclidean distance between the $i$-th predicted keypoint and its reference, and $s$ represents object scale. The parameter $\kappa_i$ is a keypoint-specific normalization constant that reflects the localization uncertainty. The visibility indicator $v_i$ specifies whether the $i$-th keypoint is visible, and $\delta(\cdot)$ denotes the indicator function. The final OKS score is obtained by aggregating the similarities over all valid keypoints. OKS-based tracking can be applied to both top-down and bottom-up models.


\subsubsection{Rep validator}
The rep validator analyzes extracted keypoints and outputs rep validity labels along with the start and end times of each rep. Algorithm \ref{alg:rep_validator} presents a high-level abstraction of the rep validation pipeline. The rep validator takes body keypoints as input. Based on the structured representation obtained through Stage 2, the validator selects movement-specific joint keypoints and computes the required joint angles and heights, thereby addressing Challenge 2 by enabling movement-aware feature extraction. These computed values form the kinematic features, which are passed to subsequent steps. Threshold parameters are used as flexible evaluation criteria across all subsequent rule-based steps.

At the beginning of a movement, the rep validator remains in an IDLE state. It monitors the kinematic features and checks whether the predefined start conditions are satisfied. If the current frame meets the start criteria, it is marked as the start of a rep. Subsequently, the movement state is set to ACTIVE, and the evaluation proceeds to the next step. Then, the rep requirement assessment is performed to determine whether the movement satisfies completion constraints by comparing with thresholds. This assessment is accumulated over the entire duration of the rep. When an end condition is detected, the algorithm records the current frame as the end time and determines its validity based on the accumulated requirement evaluation. A rep is marked as valid only if all required conditions are satisfied; otherwise, it is labeled as invalid. Upon detecting an end condition, the rep validator outputs the rep validity and corresponding start and end times, then resets to the IDLE state to await the next rep. While not explicitly shown in Algorithm~\ref{alg:rep_validator}, the judge tracks movements that do not fully satisfy the formal start or end conditions using internal state flags and subsequently categorizes such reps as invalid.

\subsubsection{Dual Strategy Cache Mechanism} 
\begin{figure}[]
  \centering
  \includegraphics[width=\linewidth]{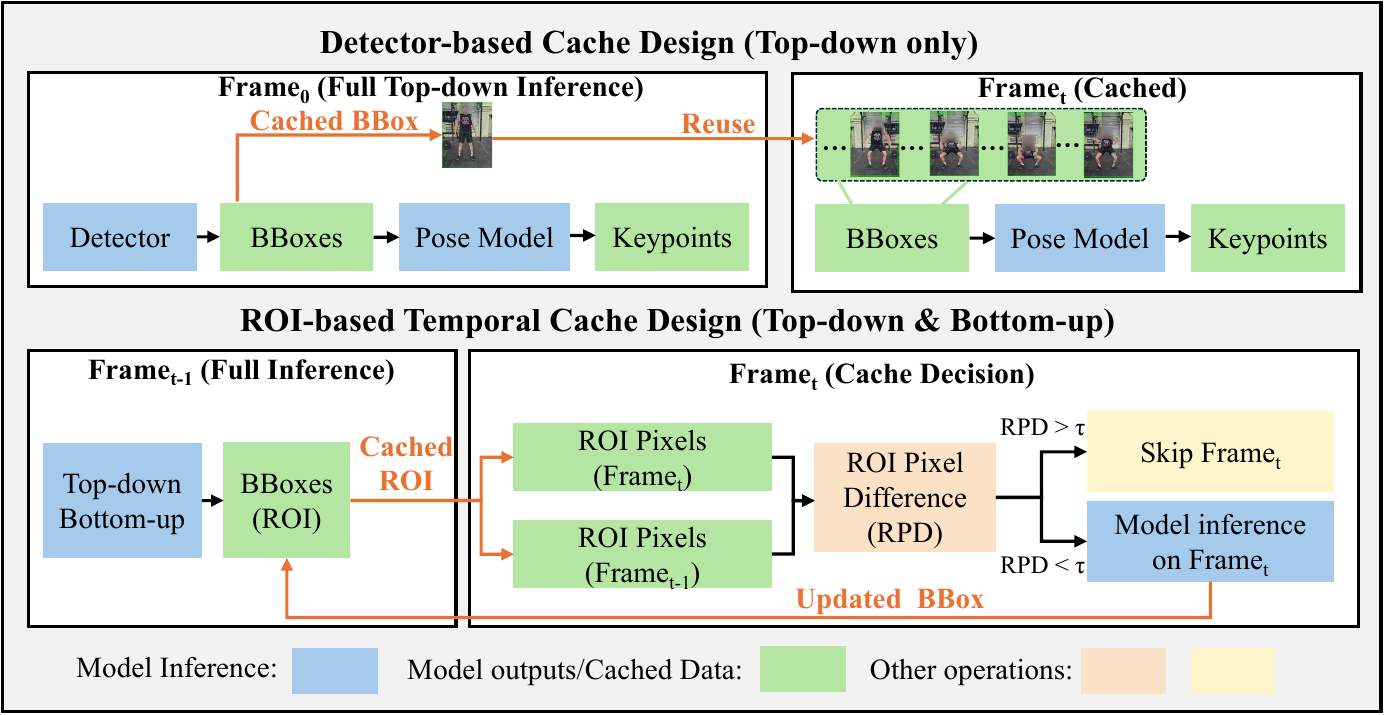}
    \caption{Overview of dual strategy cache mechanism. Frame$_0$ is the first video frame, and Frame$_t$ denotes subsequent frames.} 
  \label{fig:cache}
\end{figure}

We design a dual strategy caching  to address Challenge~3, as shown in Figure~\ref{fig:cache}.
\begin{algorithm}[t]
\tiny
\caption{Rep Validator}
\label{alg:rep_validator}
\KwIn{Keypoint sequence $\mathbf{K} = \{\mathbf{K}_t\}_{t=1}^T$, Structured representation $\mathcal{S}$, Threshold configuration $\Theta$}
\KwOut{Repetition set $\mathcal{R}$ with validity and start/end times}

\textbf{Given:} Movement judge $\mathcal{J}$, Initialize $\mathcal{R} \leftarrow \emptyset$, state $\sigma \leftarrow \texttt{IDLE}$ \\

\For{$t = 1$ \KwTo $T$}{

    Compute joint angles and heights to obtain kinematic features $\mathbf{f}_t$
    according to definitions in $\mathcal{S}$ \\
    \If{$\sigma = \texttt{IDLE}$}{
        \If{$\mathcal{J}.\text{detects\_start}(\mathbf{f}_t, \Theta)$}{
            $t_s \leftarrow t$, $\sigma \leftarrow \texttt{ACTIVE}$ \\
            
        }
    }
    \If{$\sigma = \texttt{ACTIVE}$}{
        $\mathcal{J}.\text{check\_requirements}(\mathbf{f}_t, \Theta)$ \\
    }
    \If{$\sigma = \texttt{ACTIVE}$}{
        \If{$\mathcal{J}.\text{detects\_end}(\mathbf{f}_t, \Theta)$}{
            $t_e \leftarrow t$ \\
            \If{$\mathcal{J}.\text{requirements\_satisfied}()$}{
                Append valid rep $(t_s, t_e)$ to $\mathcal{R}$
            }
            \Else{
                Append invalid rep $(t_s, t_e)$ to $\mathcal{R}$
            }
            $\sigma \leftarrow \texttt{IDLE}$ \\
        }
    }
}

\Return{$\mathcal{R}$}
\end{algorithm}
\noindent \textbf{Detector-based Cache.}
During movements, the athlete’s position remains stable with small adjustments rather than large displacements. Thus, for a top-down pose estimation pipeline, running the human detector on every frame is often redundant. After detecting the athlete in the first frame, we reuse the bbox in subsequent frames without reinvoking the detector (see top panel of Figure~\ref{fig:cache}), thereby reducing computational cost and improving efficiency. To handle small position jitters, we enlarge the bbox from the first frame by adding an extra offset. This fixed enlarged bbox covers the athlete’s possible locations across the remaining frames, ensuring that pose estimation is performed within a correct region in each frame.

\noindent \textbf{ROI-based Temporal Cache.}
\shaibal{The detector cache alone does not eliminate repeated pose inference when consecutive frames are visually similar and is applicable only to top-down models. To further reduce redundant computation beyond detector reuse, we introduce an RTC cache applicable to both top-down and bottom-up pose models.  The design assumes local temporal stability: small appearance changes within a person-centered ROI across consecutive frames indicate stable motion and allow reuse of previous results (bottom panel of Figure~\ref{fig:cache}).}

Let $F_t$ denote frame $t$. For each processed frame, we define a human ROI $R_t$ as the bbox enclosing the selected person’s predicted keypoints, enlarged by a fixed padding factor for robustness. From $R_t$, we extract a compact patch $P_t$ by grayscale conversion, Gaussian smoothing, and fixed-size downsampling. We compute an inter-frame ROI pixel difference (RPD) using Eq.~\ref{eq:rpd}.
\vspace{-2mm}
\begin{equation}
D_t = \frac{1}{N} \sum_{i=1}^{N} \left| P_t(i) - P_{t-1}(i) \right|
\label{eq:rpd}
\end{equation}
where $N$ is the number of pixels in the downsampled ROI patch.
If $D_t \le \tau$, where $\tau$ is a predefined ROI-difference threshold to skip pose inference. The cached pose keypoints from the previous frame are reused as input to the downstream rep-validation module. Otherwise, full inference runs and the ROI bbox cache are updated. The threshold $\tau$ is selected empirically per camera view via calibration sweep to maximize frame skipping while preserving rep-count correctness.

%% file: Sections/results.tex
\section{Implementation}
\subsection{Environment Setup}
\subsubsection{Datasets}\label{datasets}


To calibrate the cosine similarity threshold during the retrieval phase in Stage~1, we constructed a custom dataset due to the lack of suitable benchmarks. We used GPT-5 to generate and label context–query pairs from the IF3 and CrossFit rulebooks as relevant or irrelevant based on the rulebooks' movement rules. Labels were validated using DeepSeek R1 and human-in-the-loop review. The resulting dataset enables systematic evaluation of precision, recall, and F1 score across cosine similarity thresholds, thereby supporting optimal threshold selection during the retrieval phase in Stage~1. For movement rep evaluation, we use the CFRep dataset~\cite{alves2025repval}, an open-access benchmark for rep-based movement validation in functional fitness. It contains 64 videos of three movements captured from side, front, and diagonal views, with each rep labeled as valid (0) or invalid (1). Since annotations are sequence-level, we extend them with rep-level start and end times to enable fine-grained, rep-level evaluation.
\subsubsection{Models}
\input{Assets/Tables/pose_model}
\repjudge uses the pretrained LinQ-Mistral embedding model~\cite{choi2024linq} from HuggingFace, fine-tuned on custom dataset using AWQ quantization~\cite{lin2024awq}. GPT-4.1 is used as the generation model in Stages~1 and~2 via the OpenAI~\cite{achiam2023gpt} API. We selected GPT-4.1 after evaluating GPT-4.1, GPT-4o, GPT-4o-mini, GPT-3.5-Turbo, and DeepSeek~\cite{guo2025deepseek} in Stage~1 under identical queries and retrieved contexts. Generation quality was assessed using Gemini-2.5 Pro Flash~\cite{comanici2025gemini} as an external evaluator to avoid architectural bias, and independently by four human-in-the-loop with experience in functional fitness. Stage 3 judge system provides a wide range of pose estimation models. For evaluation, we consider pose estimation models trained on 5 different datasets, using an RTMDet-M~\cite{lyu2022rtmdet} detector for all top-down methods. In total, we evaluate 5 models covering 5 keypoint schemas, as shown in Table~\ref{tab:pose_models}. 
\subsubsection{Hardware Environments and Libraries} 
\change{We use multiple hardware environments to evaluate \repjudge. The embedding model is quantized on an NVIDIA A40 GPU (48\,GB VRAM). Source documents are pre-processed using LangChain’s PyMuPDF library\footnote{\url{https://github.com/langchain-ai/langchain}} to extract textual content. The extracted text used RedisStack\footnote{\url{https://redis.io/about/redis-stack/}} as a vector database. For fair comparison, all experiments are conducted on an NVIDIA RTX~3080 GPU. Stages~1 and~2 use Python~3.10 with PyTorch~2.5, while Stage 3 uses MMPose~1.3.2, MMCV~2.1.0, Python~3.8, and PyTorch~2.1. \shaibal{We also deploy \repjudge on a Jetson AGX Xavier edge device from NVIDIA with a TensorRT backend and evaluate system performance on both pre-recorded videos and live-streaming scenarios.}} 
\subsection{Evaluation Protocol}
\subsubsection{Evaluation metrics}
In this section, we describe the quantitative and qualitative metrics used to evaluate the effectiveness of \repjudge at different processing stages.\\
\noindent \textbf{Cosine Similarity Threshold.}
To select an appropriate cosine similarity threshold for retrieval in Stage~1, we use precision, recall, and F1 score on the dataset described in Section~\ref{datasets}. 

\noindent \textbf{Efficiency metrics.} We evaluate the embedding latency (s) and model size (GB) of the quantized embedding model relative to its full precision (FP32) baseline on NVIDIA A40 GPU. \shaibal{To characterize efficiency during rep-level judging (Stage~3), we report latency and the RTF on pre-recorded and live-streaming settings. RTF is defined as the ratio of processing time to video duration and serves as a playback-independent throughput metric that normalizes computation cost by workload size.}\\ 
\noindent\textbf{LLM and Human Evaluation Metrics.}\label{test} We evaluate generation quality using  faithfulness ($F$), completeness ($C$), and internal consistency ($S$). $F$ measures whether the structured output is supported by the provided source context without introducing hallucinated constraints. $C$ assesses whether all required movement rules and constraints are captured, while $S$ evaluates the logical coherence of the structured representation independent of rulebook correctness.  An example of calculating faithfulness is provided in Listing~\ref{lst:faithfulness_prompt}. Both $F$ and $C$ utilize a structurally similar CoT prompting strategy to ensure reasoning transparency and accuracy. These metrics are combined into a mean weighted score (MWS) using Eq.~\ref{eq:mws}:
\begin{equation}
\label{eq:mws}
\mathrm{MWS} = w_1 F + w_2 C + w_3 S
\end{equation}
In our experiment, we set $(w_1, w_2, w_3) = (0.4, 0.4, 0.2)$ to prioritize rule adherence and coverage over internal consistency.  To ensure systematic evaluation, both human annotators and LLM-based evaluators follow rubric-based scoring protocols that operationalize each metric as a structured decision procedure. Four human experts independently score each output on a 1--5 scale, with scores aggregated per movement. Scores are normalized from the 1–5 rubric scale to [0,1] before computing MWS. We analyze inter-annotator behavior using the average standard deviation (SD) and the intraclass correlation coefficient (ICC)~\cite{shrout1979intraclass}. Human and LLM evaluations are then compared using calibration and ranking measures, including the mean absolute score difference $\Delta$, the SD of per-movement human--LLM score differences, and Kendall’s $\tau$ and Spearman’s $\rho$.  
\input{Sections/prompt}
\noindent \textbf{Judge System Evaluation Metrics.}
Since our model predicts a single video containing multiple reps, directly comparing the predicted and ground-truth label sequences can lead to temporal misalignment. Such sequence-based evaluation ignores rep boundaries, allowing missed or extra reps to produce internally misaligned predictions with the same sequence length, thereby distorting evaluation metrics. To address this issue, we formulate the task as a temporal detection problem, similar to activity detection in ActivityNet~\cite{caba2015activitynet}. Our system predicts the start and end frames of each rep, defining its temporal duration, and assigns a label to each segment. For evaluation, temporal IoU (tIoU) is computed between predicted and ground-truth segments of the same class. True positives (TP), false positives (FP), and false negatives (FN) are determined based on class-consistent tIoU thresholding between predicted and ground-truth reps. In our experiments, we use a tIoU threshold of 0.2. Precision, Recall, and F1 are computed per class, and final performance is reported as the average across all classes.
\begin{figure}[]
  \centering
  \subfloat[IF3 rulebook.]{
    \includegraphics[width=0.47\linewidth]{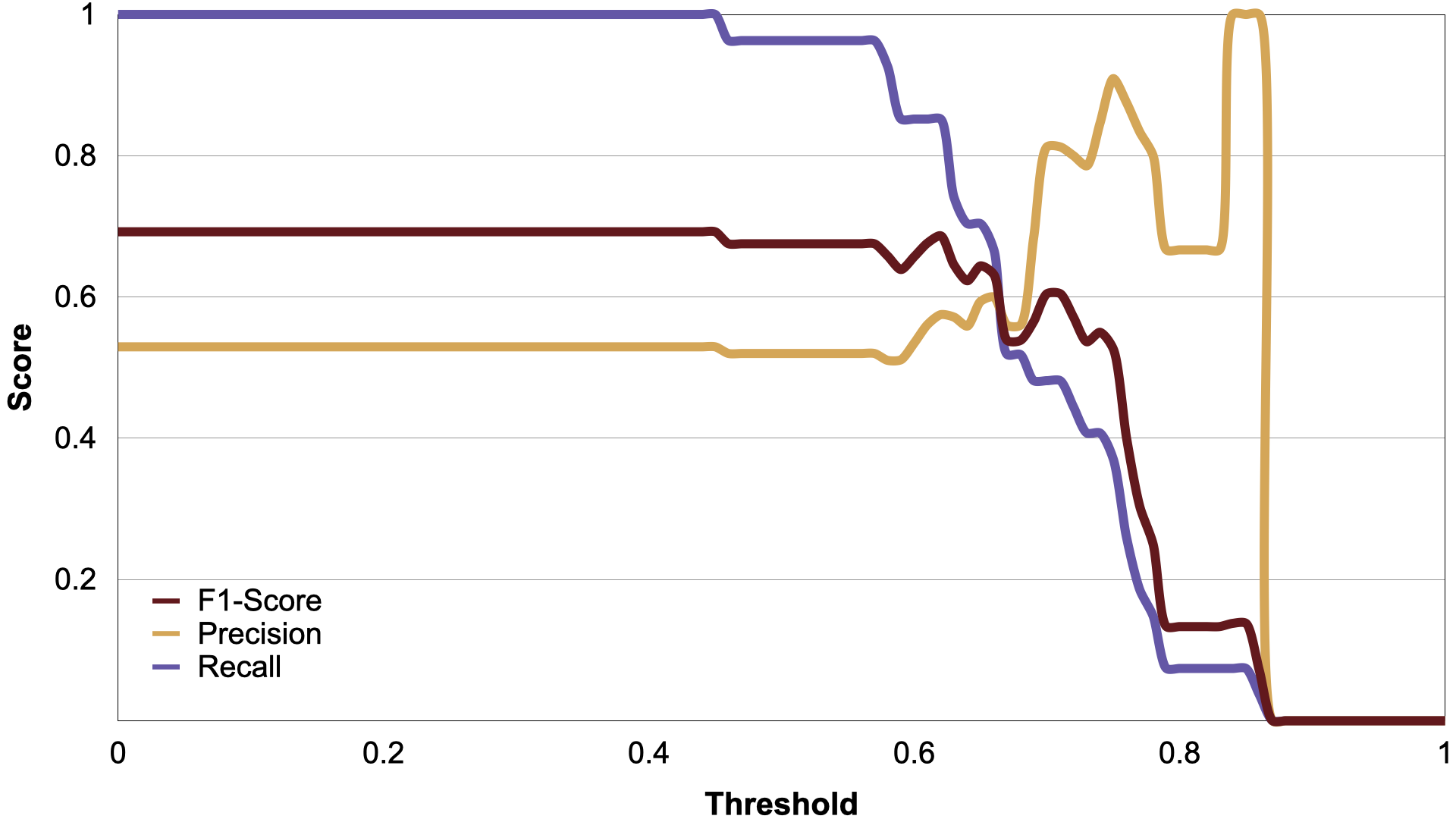}
    \label{fig:dataset_if3}
  }
  \hfill
  \subfloat[CrossFit rulebook.]{
    \includegraphics[width=0.47\linewidth]{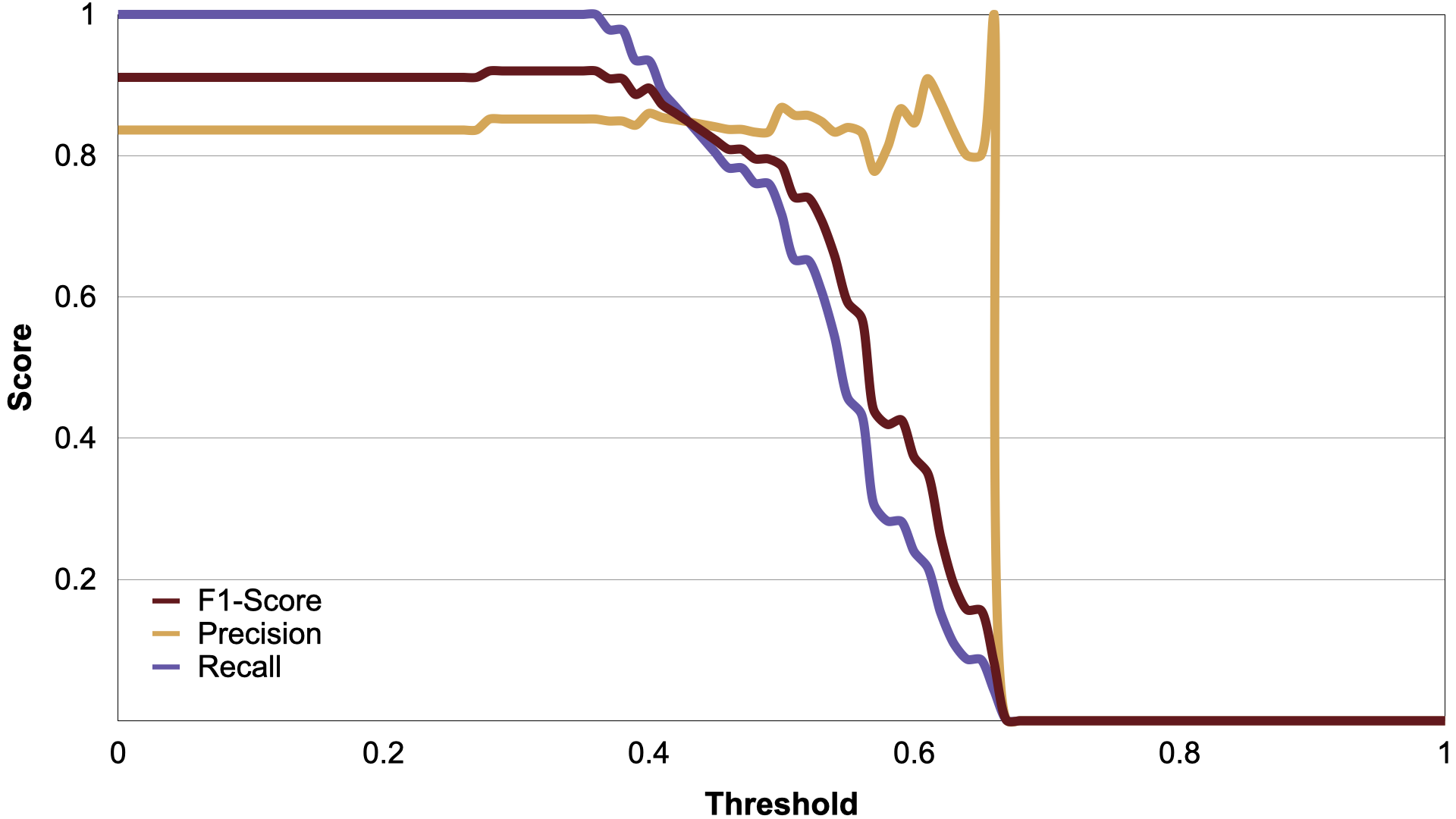}
    \label{fig:dataset_rulebook}
  }
  \caption{Cosine similarity threshold comparison for IF3 and CrossFit rulebooks.}
  \label{fig:threshold_analysis}
\end{figure}
\input{Assets/Tables/efficiency_comparison}
\subsubsection{Keypoints Schema-aware Protocol}
Pose estimation models are trained on different datasets, which define different keypoint schemas. The set of available keypoints varies across models. And the localization accuracy of keypoints may differ between models. These differences need to be considered when defining evaluation thresholds and judge functions.
\noindent \textbf{Threshold.} 
Pose estimation accuracy varies across models, and camera views introduce view-dependent prediction uncertainty; a single fixed threshold would yield inconsistent rep evaluation and unfair comparisons. In our system, thresholds are used to identify when a rep starts, is valid, and ends. To adapt the system to practical use, we perform a threshold grid search for each pose model across movements and camera views. The optimal threshold is selected by maximizing the average F1 score on the CFRep dataset.\\
\noindent\textbf{Schema-aware Judge Definition.} The available keypoints differ across pose estimation models. Therefore, we define a schema-aware judge to ensure that the evaluation criteria can be consistently applied under different keypoint schemas. For movements that require objects not directly detected by the pose models, schema-specific approximations are applied. In deadlift, since the pose models do not detect the barbell, its position is approximated using hand keypoints, with the middle finger metacarpophalangeal (MCP) joint as a proxy. For keypoint schemas without hand keypoints, the barbell position is approximated using the wrist with a fixed offset of 20, estimated as the average of the wrist–middle finger MCP distance from models that include hand keypoints. In double under, evaluation considers wrist, elbow, and shoulder angles. When hand keypoints are unavailable, wrist angles cannot be computed. In such cases, wrist angles are excluded, with only elbow and shoulder angles used for evaluation.
\input{Assets/Tables/model_evaluation_generation}
\section{\repjudge Performance Analysis}
\subsection{Embedding and Retrieval Analysis}
\noindent \textbf{Cosine Similarity Threshold Selection.}\label{threshold_analysis}
Figure~\ref{fig:threshold_analysis} compares how the precision, recall, and F1 score are affected by different cosine similarity thresholds on the custom dataset, identifying the optimal similarity cutoff for the embedding model. At lower thresholds, most context-query pairs are classified as matches, yielding high recall but lower precision due to weaker semantic matches. Precision improves as the threshold increases, but recall begins to drop as the true matches fall below the cutoff. F1 score peaks around 0.4 for the IF3 rulebook (Figure~\ref{fig:dataset_if3}), indicating a balanced trade-off between precision and recall. In contrast, the CrossFit rulebook (Figure~\ref{fig:dataset_rulebook}) achieves its optimal balance near 0.6, reflecting their textual differences. These thresholds are subsequently used during Stage~1 retrieval from the vector database to ensure contextually accurate extraction of movement standards.\\
\noindent \textbf{Efficiency Analysis for Embedding Model.}
Table~\ref{tab:embedding_efficiency} summarizes efficiency gains from quantizing the LinQ-Mistral embedding model on an NVIDIA A40 GPU. Relative to the FP32 baseline, quantization reduces model load time by $2.2\times$, embedding generation latency by $38\times$, and model size by $3.7\times$. The resulting reduction in memory footprint enables deployment of the RAG pipeline within the 10\,GB GPU memory limit of the NVIDIA RTX~3080, whereas the FP32 model cannot be loaded without memory overflow. These results demonstrate that quantization substantially improves runtime efficiency and supports practical deployment of the embedding model in RAG-based systems.
\subsection{Evaluation of LLMs in the Generation Phase} 
\noindent \textbf{Human Evaluation Results.} Table~\ref{human_evaluation_summary} summarizes human evaluation results and reliability across human experts. GPT-4.1 achieves the highest MWS (0.92), while GPT-4o and GPT-4o-mini remain competitive. In contrast, GPT-3.5-Turbo and DeepSeek obtain substantially lower MWS values, indicating weaker coverage and correctness according to the human experts. Additionally, we observe differences in agreement and discriminability across LLM models among human experts. GPT-4.1 has the lowest SD (0.30), indicating strong agreement among human experts, but a low ICC (0.07). From our observation, GPT-4.1 consistently receives high scores across movements, leaving little variance for ICC to capture. In contrast, GPT-4o-mini shows lower MWS but a higher ICC (0.61), indicating consistent agreement on relative performance differences despite lower absolute quality. GPT-4o lies between these cases, combining high MWS with moderate SD and ICC. Together, these results show that MWS reflects absolute quality, SD captures agreement, and ICC captures discriminability across movements. Those motivate our choice of GPT-4.1 by favoring consistently high-quality and accurate outputs over variability-driven discrimination, which is less critical for structuring the movement representations.\\
\input{Assets/Tables/human_llm}
\noindent\textbf{Human–LLM Calibration and Agreement.} Table~\ref{tab:human_llm} compares human and LLM evaluations using calibration and ranking measures. The absolute difference between mean human and LLM scores ($\Delta$) remains small across all models (0.00–0.06), indicating limited calibration bias. GPT-4o shows the smallest difference, while GPT-4.1 shows a slightly larger gap, indicating minor underestimation by the LLM evaluator. Beyond mean calibration, we examine how score differences vary across movements for human and LLM evaluation. GPT-4.1 exhibits the lowest dispersion, indicating consistently small SD from human evaluation, while other models show higher variability.
We further evaluate rank-based consistency using Kendall’s $\tau$ and Spearman’s $\rho$. GPT-4o shows the strongest agreement with human rankings, achieving the highest $\tau$ (0.84) and $\rho$ (0.88), indicating stable ordering across movements. GPT-4.1 also exhibits strong alignment\mycut{($\tau = 0.72$, $\rho = 0.79$)}, while GPT-4o-mini follows with slightly lower but comparable agreement. In contrast, GPT-3.5-mini and DeepSeek exhibit weaker rank alignment, indicating more frequent discrepancies relative to human judgments. Overall, the strong calibration and rank agreement between human and LLM evaluator indicate that LLM-based assessment can reliably support scalable validation of rule-structuring quality in \repjudge.\\
\noindent \textbf{Model Selection Rationale.} Based on our experimental results, we select GPT-4.1 as the generation model due to its higher human-rated quality. GPT-4.1 attains the highest MWS, the lowest annotator disagreement, and maintains strong alignment with both human and LLM evaluator. Its lower ICC is mainly due to score concentration and does not indicate instability.
\input{Assets/Tables/judge_system}
\vspace{-.3mm}
\subsection{Judge System Performance Evaluation}
Tables \ref{tab:squat_results}, \ref{tab:deadlift_results}, and \ref{tab:double_unders_results} report the performance of squat, deadlift, and double under across pose models and camera views on the CFRep dataset without caching. Precision, recall, and F1 are averaged across classes. From the results, we observe that camera view has a significant impact on the performance of the rep judge system. This is because different camera views affect pose observability and keypoint estimation, impacting rule-based evaluation. However, an unobstructed front view does not always yield better results, as it introduces deviations in certain joint angle calculations. When evaluating upright body alignment, a front view provides limited depth information, making it difficult to capture slight forward or backward torso inclination due to perspective foreshortening. 

As shown in Table~\ref{tab:deadlift_results}, the front view performs worse for deadlift, which mainly evaluates back inclination along the depth direction. In contrast, the side view yields the best performance because it clearly reveals depth-related posture changes. For squat (Table~\ref{tab:squat_results}), the front view gives the best performance, as the main motion involves vertical movement of the center of mass, with clear upward and downward motion of the hips and shoulders along the y-axis. For double under (Table~\ref{tab:double_unders_results}), the side view achieves the best performance, since it selects the side with higher keypoint confidence for evaluation, and provides more reliable observation of one body side. Among the three movements, squat achieves the highest average F1 score with HRNet-W32 (0.90), while RTMPose-L performs best on deadlift (0.89) and double under with RTMPose-X (0.83). Different pose models use different architectures and training data. These differences affect how well each model captures movement geometry under different viewpoints. As a result, performance varies across movements and camera views. In addition, these best-performing models show relatively balanced precision and recall, indicating similar performance across the two classes. However, double under shows lower performance than the other two movements. This is mainly because double under requires evaluating two arm rotations within a single jump. Since the dataset videos are recorded at 30 frames per second, the fast arm motion in double under cannot be sufficiently captured, resulting in limited observable information for accurate evaluation. Overall, under the optimal camera view for each movement, \repjudge demonstrates reliable performance on rep validation.
\subsection{Dual Strategy Cache Performance}
\shaibal{Table~\ref{cacheacc} evaluates the impact of different cache strategies on performance. We select one top-down model (RTMPose-L) and one bottom-up model (RTMO-L). In the top-down model, two cache strategies and their combination are applied. The bottom-up model does not include a detection model, only the RTC cache can be applied. As cache strategies introduce differences in the input to the judge system compared to the full inference pipeline, slight performance changes are expected.
From the table, we observe that two cache strategies and their combined version have only a slight impact on performance. In the top-down model, the DC achieves the best performance on the squat, with the highest average performance change over three views (+0.05). RTC caching performs slightly better on deadlift and double under, with average changes of +0.02 and -0.01, respectively.
In the bottom-up model, RTC achieves good performance on the squat and deadlift but shows a drop on double under. This is because double under is more challenging. In the squat and deadlift, the person’s position relative to the ground is largely stable, whereas in double under, the movement is more variable and complex. As a result, the ROI location varies more significantly, potentially reducing the effectiveness of RTC.
Overall, the proposed cache strategies accelerate inference with minimal performance impact. In addition, the RTC caching can be applied to different pose models, demonstrating the flexibility of our design.}


\begin{figure*}[]
\centering

{\footnotesize
\textcolor[HTML]{AEB4BB}{\raisebox{-0.3ex}{\rule{7pt}{7pt}}}~W/o Cache \quad\quad
\textcolor[HTML]{F3DCA4}{\raisebox{-0.3ex}{\rule{7pt}{7pt}}}~DC \quad\quad
\textcolor[HTML]{E8C06D}{\raisebox{-0.3ex}{\rule{7pt}{7pt}}}~RTC \quad\quad
\textcolor[HTML]{C99E3F}{\raisebox{-0.3ex}{\rule{7pt}{7pt}}}~Combined
}

\vspace{2pt}

\subfloat[RTX 3080 (pre-recorded)\label{latencygpu}]{
    \includegraphics[width=0.31\linewidth,height=2.54cm]{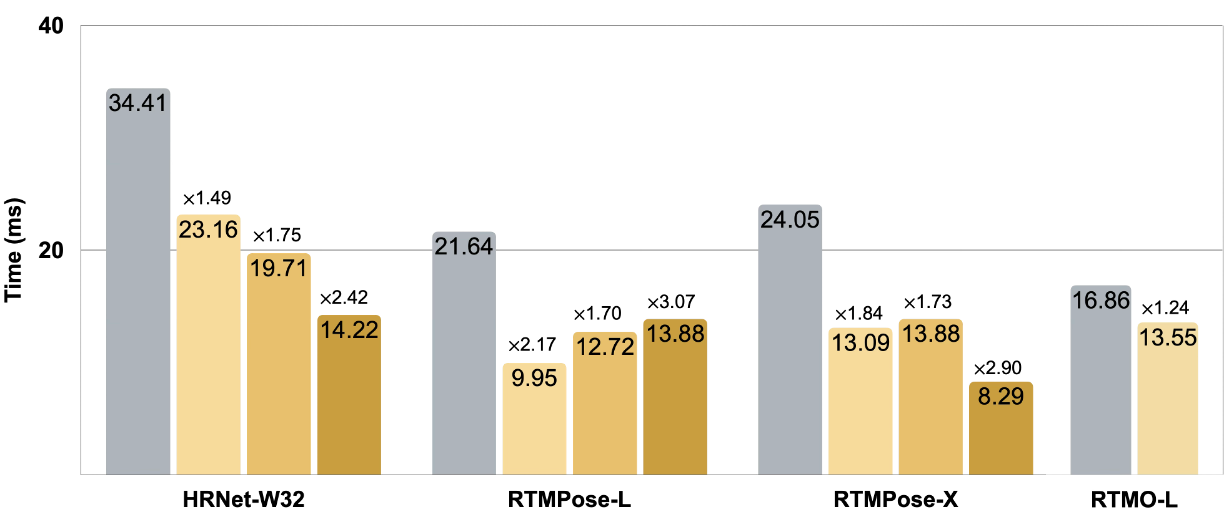}
}
\hfill
\subfloat[Edge (pre-recorded)\label{prerecor}]{
    \includegraphics[width=0.32\linewidth]{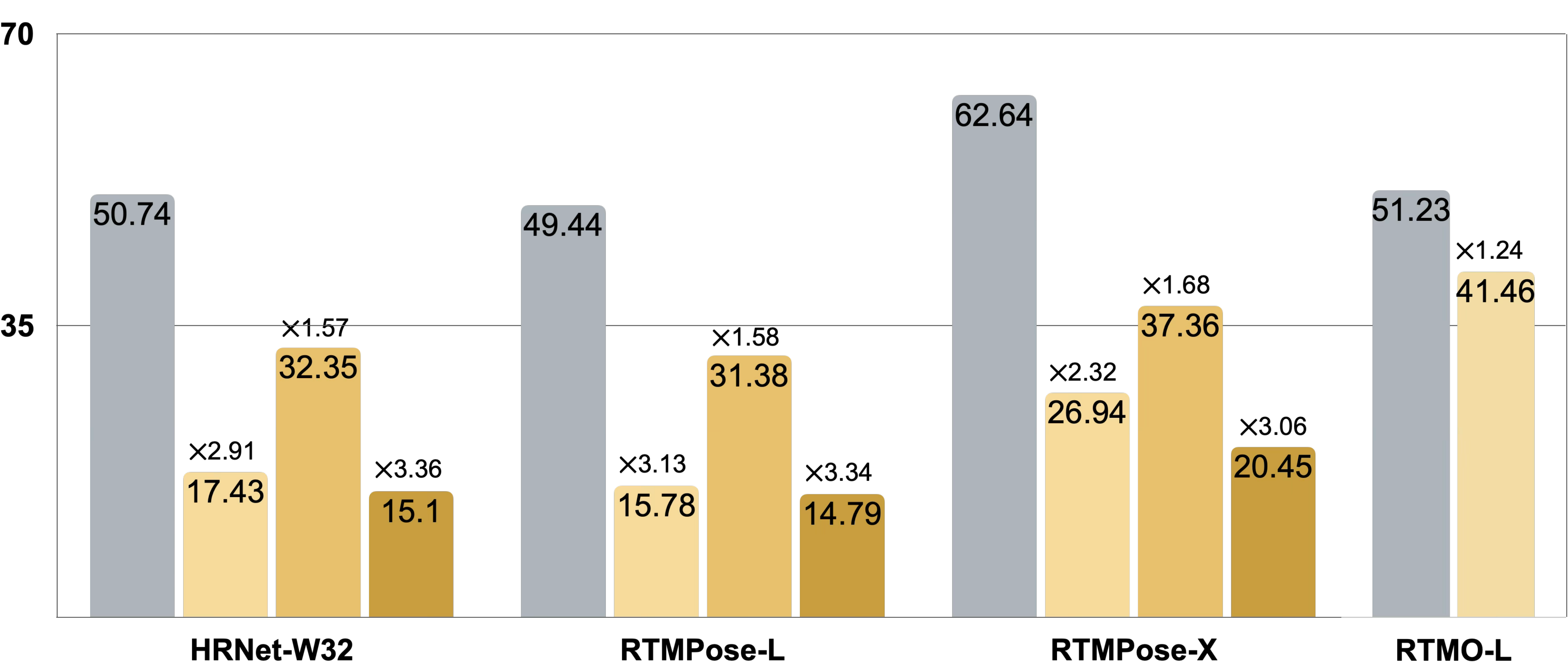}
}
\hfill
\subfloat[Edge (live-streaming)\label{livestream}]{
    \includegraphics[width=0.31\linewidth]{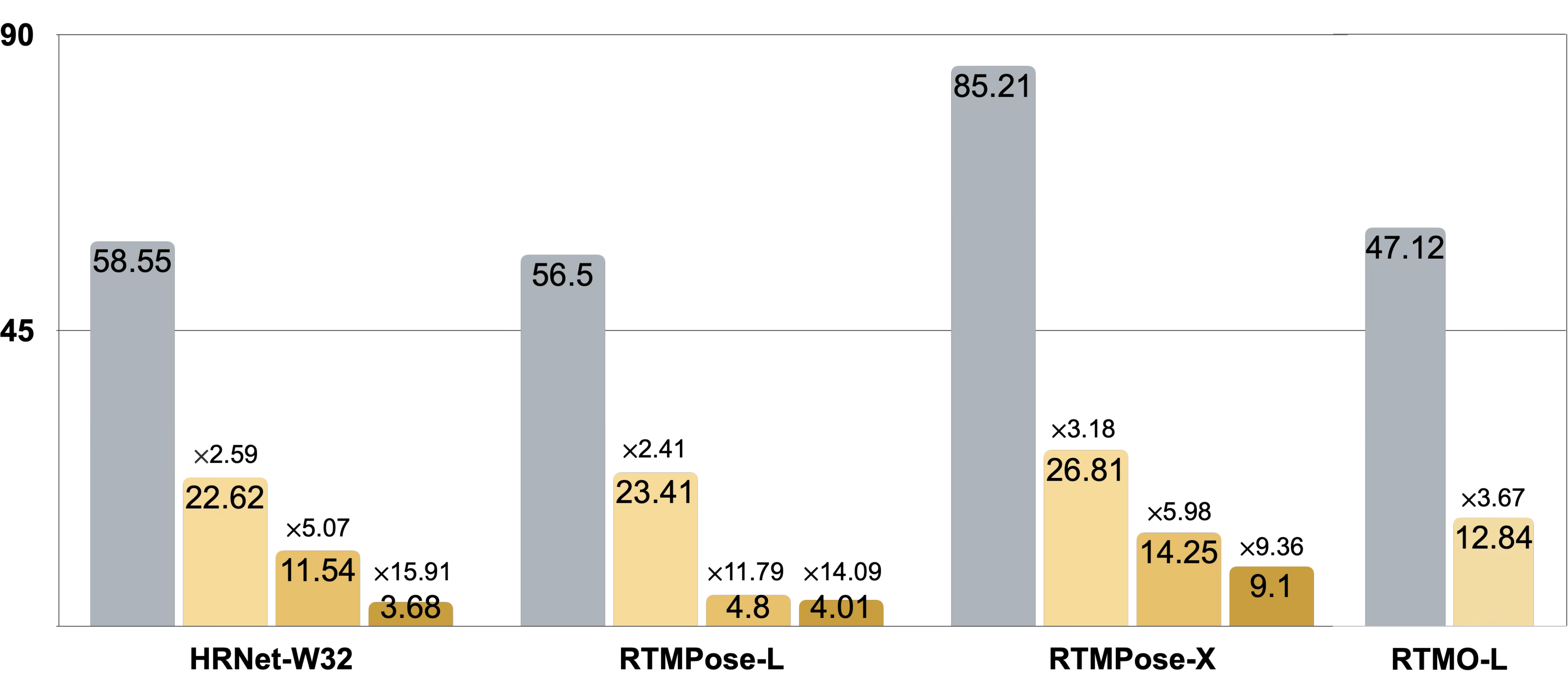}
}

\caption{Rep-level latency and speedup ($\times$ relative to the w/o cache baseline) across different pose estimation models using cache-based techniques on edge devices.}
\label{fig:latency_all}

\vspace{-0.5em}
\end{figure*}
\input{Assets/Tables/cache_performance}

\subsection{Efficiency on Edge Devices}
\shaibal{Since Stages~1 and~2 are performed offline per movement definition, efficiency analysis focuses on Stage~3, which dominates per-video inference cost and practical judging throughput. Figures~\ref{fig:latency_all} and~\ref{fig:rtf_and_adapt} summarize rep-level latency and RTF for \repjudge on edge devices, including the high-performance RTX~3080 and the resource-constrained AGX Xavier, under both pre-recorded and live-streaming settings. Here, rep-level latency is measured at the decision frame where a rep is classified as valid or invalid.}
\subsubsection{Evaluation on RTX 3080} \shaibal{As illustrated in Figure~\ref{latencygpu}, \repjudge achieves substantial latency reductions across top-down and bottom-up pose estimation models on RTX~3080. For example, RTMPose-X reduces average rep-level latency from $24.05$ ms to $13.88$ ms with RTC ($1.73\times$ speedup). With the combined cache, latency further drops to $8.29$ ms, corresponding to a $2.9\times$ speedup. Beyond latency, Figure~\ref{fig:rtf_3080} reports RTF across pose estimation models on a 10\-s video using RTX 3080. Several models already achieve RTF below 1.0 at baseline, while the heatmap-based HRNet-W32 operates close to the input-rate requirement. With caching techniques, HRNet-W32 achieves a $2.42\times$ speedup, reducing its RTF to a stable $0.504$ and ensuring reliable system performance.}

\subsubsection{Evaluation on Jetson AGX Xavier} We evaluate \repjudge efficiency for pre-recorded and live-streaming videos.

\begin{figure}[t]
\centering
\footnotesize
\begin{tikzpicture}
\node[
draw=gray!60,
fill=white,
line width=0.15pt,
rounded corners=1.5pt,
inner xsep=2pt,
inner ysep=2pt
] {\tiny
\legboxgray\,W/o Cache\hspace{3pt}
\legbox{crosshatch}\,DC\hspace{3pt}
\legbox{vertical lines}\,RTC\hspace{3pt}
\tikz\draw[fill=cachegold!75,draw=black,line width=0.25pt]
(0,0) rectangle (0.24,0.15);\,Combined\hspace{3pt}
\tikz[baseline=-0.6ex]{%
\draw[dash pattern=on 4pt off 2pt,line width=0.8pt,color=baselinegray]
(0,0)--(0.36,0);
}\!RTF=1
};
\end{tikzpicture}

\vspace{2pt}

\subfloat[RTX 3080\label{fig:rtf_3080}]{
  \includegraphics[width=0.47\linewidth]{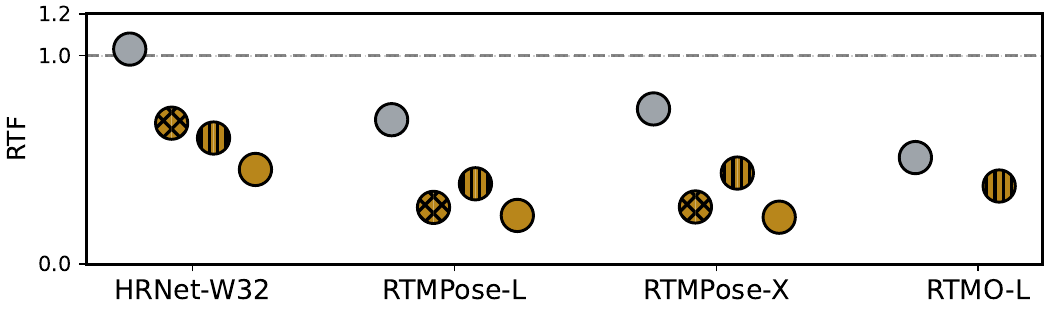}
}
\hfill
\subfloat[AGX Xavier\label{fig:rtf_AGX}]{
  \includegraphics[width=0.47\linewidth]{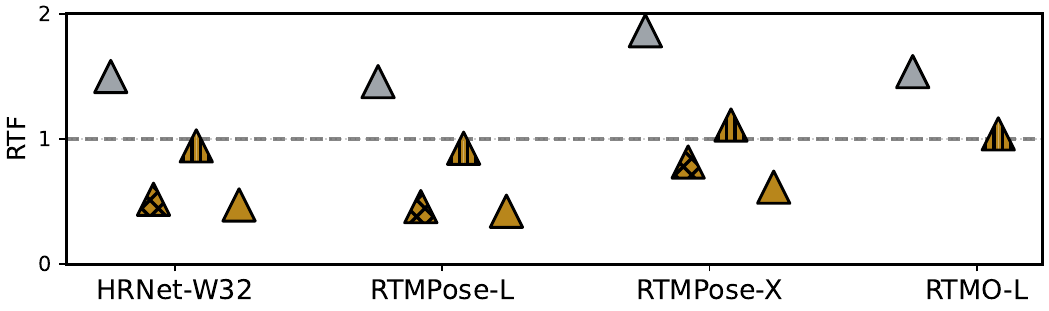}
}

\caption{RTF analysis across pose models with caching strategies on a 10-s video (300 frames) on edge devices.}
\label{fig:rtf_and_adapt}

\vspace{-0.5em}
\end{figure}
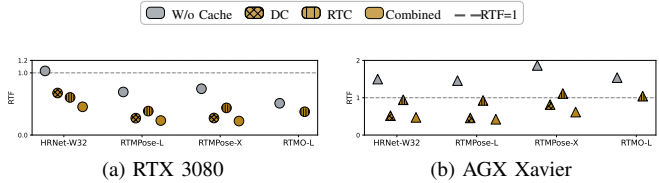

\noindent\textbf{Pre-recorded Settings.} As illustrated in Figure~\ref{prerecor}, rep-level latency drops significantly with caching on AGX Xavier, especially for heavier backbones. For example, HRNet-W32 decreases from 50.74\,ms to 15.10\,ms with the combined cache strategy, yielding up to a $3.36\times$ speedup. RTMPose-X shows a similar trend, achieving up to $3.06\times$ speedup. These latency reductions are also reflected in the corresponding RTF values. On AGX Xavier (Figure~\ref{fig:rtf_AGX}), baseline configurations show RTF above 1 on the 10\-s video. This is because the detector and pose inference run on every frame, which increases the computational load on the edge device. However, DC and RTC caching mitigate this overhead when enabled (RTF $<$ 1), allowing processing faster than the input rate on AGX Xavier.\\
\noindent\textbf{Live-streaming.} In live-streaming (Figure~\ref{livestream}), caching enables substantial speedups across models. HRNet-W32 latency decreases from 58.55\,ms to 3.68\,ms with the combined cache configuration ($15.91\times$ speedup), while RTMPose-L reduces from 56.5\,ms to 4.01\,ms ($14.09\times$).  We further observe that RTC provides a greater speedup than DC when applied individually. This is perhaps due to stronger regional-level continuity between consecutive live-streaming frames, even when detector bounding boxes fluctuate due to natural motion. As a result, RTC allows more frames to skip full pose inference safely, leading to higher efficiency gains. 

\begin{figure}[t]
\centering
\subfloat[\label{fig:yeartoyear}]{
  \includegraphics[width=0.59\linewidth]{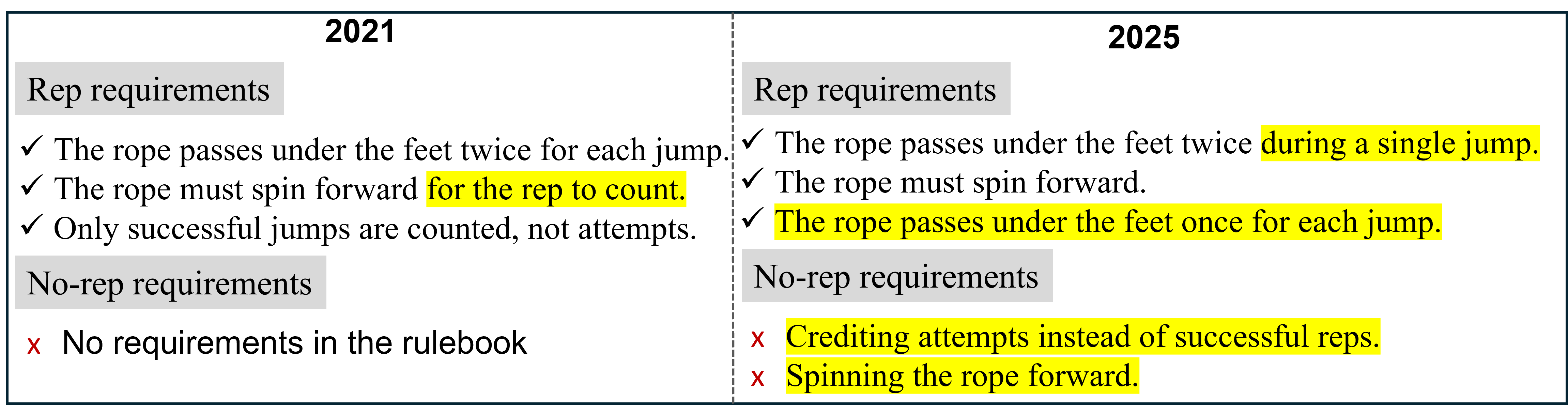}
}
\hfill
\subfloat[\label{fig:adapt}]{
  \includegraphics[width=0.35\linewidth]{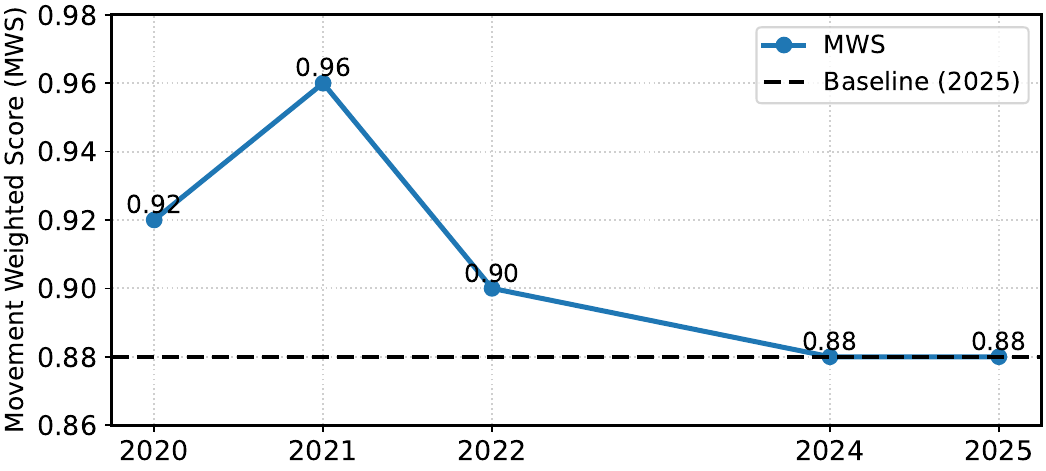}
}

\caption{Adaptability analysis of \repjudge on double under. (a) Rulebook revisions between years with modified clauses highlighted. (b) Adaptability over the years across CrossFit competitions measured by normalized MWS.}
\label{fig:adapt_year}

\vspace{-0.5em}
\end{figure}
\subsubsection{Discussion} As shown in Figs.~\ref{fig:latency_all} and~\ref{fig:rtf_and_adapt}, \repjudge achieves efficiency gains on edge devices. Rep-level judgment can be done well within practical judging requirements for both pre-recorded and live-streaming scenarios. While human judges rely on sequential observation, with a visual reaction time of 200–250\,ms~\cite{jain2015comparative}, fast fitness movements often contain critical validity cues (e.g., depth or joint lockout) that persist for only a few frames. This creates a high perceptual load and increases the risk of inconsistent calls.

In contrast, our results show that \repjudge consistently maintains low rep-level latency and a low RTF. By achieving an RTF below 1.0, the system processes motion faster than real-time playback. This ensures no visual evidence is dropped even during high-speed movements. Furthermore, this throughput enables the system to evaluate multiple video streams in parallel on available compute resources, supporting scalable, asynchronous analysis. Ultimately, \repjudge overcomes the temporal constraints of manual observation, serving as a practical tool that assists human judges by providing a consistent, reliable assessment of fitness movements.
\subsection{Adaptability Analysis} We use MWS to measure Stage~1–2 adaptability across different CrossFit competitions for the double under movement. Figure~\ref{fig:yeartoyear} shows rulebook differences between the 2021 and 2025 versions. The newer rulebook adds additional no-rep conditions and more explicit constraint clauses. Despite these added constraints, Stage~1–2 evaluation remains stable. Figure~\ref{fig:adapt} shows the normalized MWS across rulebook variants relative to the baseline\footnote{\href{https://games.crossfit.com/workouts/open/2025/2\#movementStandards}{2025 CrossFit Competition}}. The MWS values remain high across rulebooks, indicating consistent rule retrieval and structuring under rule variation. A small MWS decrease from 2021 to 2025 is observed due to the added no-rep conditions. These results confirm the adaptability of the pipeline, allowing rule standards to be updated through Stage~1–2 processing without changing the underlying system architecture.

%% file: Assets/Tables/pose_model.tex
\begin{table}[]
\centering
\footnotesize
\setlength{\tabcolsep}{3pt}
\caption{Pose models used in \repjudge for evaluation. Model names correspond to configuration file names in the MMPose.
Models above the divider are \emph{top-down} methods, while models below are \emph{bottom-up} methods.}
\label{tab:pose_models}
\begin{tabular}{@{}l l l@{}}
\toprule
\textbf{Model Name} & \textbf{Config Name} & \textbf{Data Type} \\
\midrule
HRNet-W32 & \mbox{td-hm\_hrnet-w32\_8xb64\_aic-256x192} & AIC \\
RTMPose-L & \mbox{rtmpose-l\_8xb256\_humanart-256x192} & HumanArt \\
RTMPose-X & \mbox{rtmpose-x\_8xb256\_body8-halpe26-384x288} & HALPE-26 \\
\midrule
RTMO-L & \mbox{rtmo-l\_16xb16\_body7-640x640} & Body7 \\
DEKR & \mbox{dekr\_hrnet-w48\_8xb5\_crowdpose-640x640} & CrowdPose \\
\bottomrule
\end{tabular}
\end{table}


%% file: Sections/prompt.tex
\begin{listing}[t]
\tiny
\vspace{-0.5em}
\caption{Faithfulness Evaluation Protocol.}
\label{lst:faithfulness_prompt}
\centering

\begin{tcolorbox}[
  enhanced,
  breakable,
  boxrule=0.6pt,
  colframe=black!30,
  colback=white,
  arc=0pt,
  boxsep=0pt,
  left=0pt,
  right=0pt,
  top=0pt,
  bottom=0pt
]

\begin{tcolorbox}[
  colback=blue!2!white,
  boxrule=0pt,
  sharp corners,
  before skip=0pt,
  after skip=0pt,
  boxsep=0pt,
  top=0pt,
  bottom=0pt
]
\begin{lstlisting}[style=promptstyle]
Role: You are an expert evaluator judging FAITHFULNESS of a movement extraction.
Task: Judge the FAITHFULNESS of a movement extraction.
\end{lstlisting}
\end{tcolorbox}

\begin{tcolorbox}[
  colback=cyan!2!white,
  boxrule=0pt,
  sharp corners,
  before skip=0pt,
  after skip=0pt,
  boxsep=0pt,
  top=0pt,
  bottom=0pt
]
\begin{lstlisting}[style=promptstyle]
DEFINITION
Faithfulness evaluates whether the candidate JSON contains information supported by the provided source material.
(Missing information is NOT penalized.)
A faithful extraction:
- Introduces no unsupported constraints
- Hallucinates no new rules or requirements
- May omit valid but unused information
\end{lstlisting}
\end{tcolorbox}

\begin{tcolorbox}[
  colback=green!2!white,
  boxrule=0pt,
  sharp corners,
  before skip=0pt,
  after skip=0pt,
  boxsep=0pt,
  top=0pt,
  bottom=0pt
]
\begin{lstlisting}[style=promptstyle]
INPUTS
Reference - Movement Rules: {movement rules}
Reference - Definition for extended rules: {definitions}
Candidate - Structured JSON: {response from LLM}
\end{lstlisting}
\end{tcolorbox}

\begin{tcolorbox}[
  colback=orange!2!white,
  boxrule=0pt,
  sharp corners,
  before skip=0pt,
  after skip=0pt,
  boxsep=0pt,
  top=2pt,
  bottom=2pt
]
\begin{lstlisting}[style=promptstyle]
EVALUATION PROCEDURE
Step 1: Identify Claims
- Enumerate each distinct constraint or requirement expressed in the JSON.
Step 2: Verify Against Source
For each claim:
- Explicitly stated in reference -> SUPPORTED
- Reasonably inferred from reference -> SUPPORTED
- Not supported or contradictory -> UNSUPPORTED
Step 3: Hallucination Classification
- MINOR: Plausible but weakly supported inference
- MAJOR: Completely unsupported or contradictory claim
Step 4: Score Assignment
5 = All claims supported
4 = Supported with minor reasonable inferences
3 = Minor unsupported claims
2 = Major or multiple unsupported claims
1 = Many unsupported claims
0 = Mostly or entirely hallucinated
Step 5: Self-Consistency Check
- Ensure rubric alignment
- Do NOT penalize missing information
- Allow reasonable inference
\end{lstlisting}
\end{tcolorbox}

\begin{tcolorbox}[
  colback=blue!2!white,
  boxrule=0pt,
  sharp corners,
  before skip=0pt,
  after skip=0pt,
  boxsep=0pt,
  top=0pt,
  bottom=0pt
]
\begin{lstlisting}[style=promptstyle]
Decision: Provide a brief justification and SCORE: <0--5>
\end{lstlisting}
\end{tcolorbox}

\end{tcolorbox}
\vspace{-0.4em}
\end{listing}

%% file: Assets/Tables/efficiency_comparison.tex
\begin{table}[]
\renewcommand{\arraystretch}{0.9}
\centering
\caption{Efficiency comparison of embedding model with and without quantization (QEM) on NVIDIA A40.}
\label{tab:embedding_efficiency}
\begin{tabular}{lccc}
\hline
\textbf{Metrics} & \textbf{FP32} & \textbf{QEM} & \textbf{Improvement} \\
\hline
Load time (s)        & 12.29 & 5.68  & $\downarrow\,2.2\times$ \\
Latency (s)     & 1.07  & 0.028 & $\downarrow\,38\times$ \\
Model size (GB)           & 13.27 & 3.61  & $\downarrow\,3.7\times$ \\
\hline
\end{tabular}
\end{table}

%% file: Assets/Tables/model_evaluation_generation.tex
\begin{table}[]
\centering
\renewcommand{\arraystretch}{0.9}
\caption{Summary of human–human evaluation. MWS denotes the weighted movement-averaged human score. SD reports within-movement variability, and ICC measures consistency across movements.}
\label{human_evaluation_summary}
\begin{tabular}{lccc}
\hline
\textbf{Model}         & \textbf{MWS $\uparrow$} & \textbf{SD $\downarrow$} & \textbf{ICC $\uparrow$} \\ \hline
GPT-4.1      & \textbf{0.92}                          & \textbf{0.30}                     & 0.07                   \\ 
GPT-4o        & 0.82                          & 0.44                     & 0.31                   \\
GPT-4o-mini   & 0.75                          & 0.37                     & \textbf{0.61}                  \\ 
GPT-3.5-Turbo & 0.55                          & 0.61                     & 0.41               \\    
Deepseek      & 0.592                          & 0.64                     & 0.13                    \\ \hline
\end{tabular}
\end{table}

%% file: Assets/Tables/human_llm.tex
\begin{table}[]
\scriptsize
\renewcommand{\arraystretch}{0.85}
\caption{Human–LLM calibration and ranking consistency across movements. $\Delta$ is the absolute difference between mean human and LLM scores, and SD is the standard deviation of per-movement differences. Kendall’s $\tau$ and Spearman’s $\rho$ measure ranking consistency.}
\label{tab:human_llm}
\resizebox{\columnwidth}{!}{%
\begin{tabular}{lccclcc}
\hline
\textbf{Model}        & \multicolumn{1}{l}{\textbf{Human}} & \multicolumn{1}{l}{\textbf{LLM}} & \multicolumn{1}{l}{\textbf{$\Delta$}} & \multicolumn{1}{c}{\textbf{SD $\downarrow$}} & \multicolumn{1}{l}{\textbf{$\boldsymbol \tau$ $\uparrow$}} & \multicolumn{1}{l}{\textbf{$\boldsymbol \rho$ $\uparrow$}} \\ \hline
GPT-4.1      & 0.92                               & 0.86                             & 0.06                                  & 0.03                                         & 0.72                                         & 0.79                                         \\
GPT-4o       & 0.82                               & 0.83                             & 0.01                                  & 0.05                                         & 0.84                                         & 0.88                                         \\
GPT-4o-mini  & 0.75                               & 0.72                             & 0.03                                  & 0.06                                         & 0.72                                         & 0.75                                         \\
GPT-3.5-Turbo & 0.52                               & 0.52                             & 0.00                                  & 0.11                                         & 0.55                                         & 0.55                                         \\
DeepSeek     & 0.59                               & 0.57                             & 0.02                                  & 0.05                                         & 0.57                                         & 0.61                                         \\ \hline
\end{tabular}%
}
\end{table}

%% file: Assets/Tables/judge_system.tex
\begin{table}[t]
\scriptsize
\centering
\renewcommand{\arraystretch}{0.9}
\setlength{\aboverulesep}{0.3ex}
\setlength{\belowrulesep}{0.3ex}
\setlength{\cmidrulesep}{0.2ex}
\caption{Performance comparison for Squat under different camera views and pose models.}
\label{tab:squat_results}
\resizebox{\columnwidth}{!}{
\begin{tabular}{llccc}
\toprule
\textbf{Model Name} & \textbf{View} &
\textbf{Avg Precision} & \textbf{Avg Recall} & \textbf{Avg F1} \\
\midrule

\multirow{3}{*}{HRNet-W32}
& Front & 0.89 & 0.91 & \textbf{0.90} \\
& Diag  & 0.78 & 0.60 & 0.57 \\
& Side  & 0.95 & 0.65 & 0.71 \\
\cmidrule{1-5}

\multirow{3}{*}{RTMPose-L}
& Front & 0.93 & 0.86 & \textbf{0.89} \\
& Diag  & 0.80 & 0.63 & 0.62 \\
& Side  & 0.95 & 0.65 & 0.71 \\
\cmidrule{1-5}

\multirow{3}{*}{RTMPose-X}
& Front & 0.91 & 0.88 & \textbf{0.89} \\
& Diag  & 0.80 & 0.63 & 0.62 \\
& Side  & 0.95 & 0.65 & 0.71 \\
\specialrule{0.8pt}{0pt}{0pt}

\multirow{3}{*}{RTMO-L}
& Front & 0.91 & 0.88 & \textbf{0.89} \\
& Diag  & 0.78 & 0.65 & 0.65 \\
& Side  & 0.95 & 0.65 & 0.71 \\
\cmidrule{1-5}

\multirow{3}{*}{DEKR}
& Front & 0.78 & 0.79 & \textbf{0.79} \\
& Diag  & 0.71 & 0.58 & 0.54 \\
& Side  & 0.95 & 0.64 & 0.70 \\

\bottomrule
\end{tabular}
}
\end{table}
\begin{table}[t]
\scriptsize
\centering
\renewcommand{\arraystretch}{0.9}
\setlength{\aboverulesep}{0.3ex}
\setlength{\belowrulesep}{0.3ex}
\setlength{\cmidrulesep}{0.2ex}
\caption{Performance comparison for Deadlift under different camera views and pose models.}
\label{tab:deadlift_results}
\resizebox{\columnwidth}{!}{
\begin{tabular}{llccc}
\toprule
\textbf{Model Name} & \textbf{View} &
\textbf{Avg Precision} & \textbf{Avg Recall} & \textbf{Avg F1} \\
\midrule
\multirow{3}{*}{HRNet-W32}
& Front & 0.64 & 0.47 & 0.54 \\
& Diag  & 0.87 & 0.66 & 0.67 \\
& Side  & 0.87 & 0.67 & \textbf{0.74} \\
\cmidrule{1-5}

\multirow{3}{*}{RTMPose-L}
& Front & 0.62 & 0.53 & 0.55 \\
& Diag  & 0.84 & 0.57 & 0.55 \\
& Side  & 0.96 & 0.86 & \textbf{0.89} \\
\cmidrule{1-5}

\multirow{3}{*}{RTMPose-X}
& Front & 0.45 & 0.49 & 0.47 \\
& Diag  & 0.83 & 0.50 & 0.42 \\
& Side  & 0.89 & 0.55 & \textbf{0.52} \\
\specialrule{0.8pt}{0pt}{0pt}

\multirow{3}{*}{RTMO-L}
& Front & 0.70 & 0.55 & 0.57 \\
& Diag  & 0.87 & 0.66 & 0.67 \\
& Side  & 0.84 & 0.80 & \textbf{0.82} \\
\cmidrule{1-5}

\multirow{3}{*}{DEKR}
& Front & 0.57 & 0.66 & 0.53 \\
& Diag  & 0.58 & 0.64 & \textbf{0.61} \\
& Side  & 0.56 & 0.63 & 0.59 \\
\bottomrule
\end{tabular}
}
\end{table}

\begin{table}[t]
\scriptsize
\centering
\renewcommand{\arraystretch}{0.9}
\setlength{\aboverulesep}{0.3ex}
\setlength{\belowrulesep}{0.3ex}
\setlength{\cmidrulesep}{0.2ex}
\caption{Performance comparison for Double under across different camera views and pose models.}
\label{tab:double_unders_results}
\resizebox{\columnwidth}{!}{
\begin{tabular}{llccc}
\toprule
\textbf{Model Name} & \textbf{View} &
\textbf{Avg Precision} & \textbf{Avg Recall} & \textbf{Avg F1} \\
\midrule
\multirow{3}{*}{HRNet-W32}
& Front & 0.53 & 0.53 & 0.49 \\
& Diag  & 0.73 & 0.63 & 0.63 \\
& Side & 0.76 & 0.75 & \textbf{0.75} \\
\cmidrule{1-5}

\multirow{3}{*}{RTMPose-L}
& Front & 0.60 & 0.63 & 0.45 \\
& Diag  & 0.64 & 0.62 & 0.62 \\
& Side  & 0.80 & 0.81 & \textbf{0.80} \\
\cmidrule{1-5}

\multirow{3}{*}{RTMPose-X}
& Front & 0.57 & 0.60 & 0.50 \\
& Diag  & 0.71 & 0.68 & 0.69 \\
& Side  & 0.82 & 0.84 & \textbf{0.83} \\
\specialrule{0.8pt}{0pt}{0pt}

\multirow{3}{*}{RTMO-L}
& Front & 0.54 & 0.56 & 0.47 \\
& Diag  & 0.60 & 0.58 & 0.58 \\
& Side  & 0.88 & 0.74 & \textbf{0.79} \\
\cmidrule{1-5}

\multirow{3}{*}{DEKR}
& Front & 0.57 & 0.62 & 0.55 \\
& Diag  & 0.77 & 0.58 & 0.58 \\
& Side  & 0.73 & 0.71 & \textbf{0.72} \\
\bottomrule
\end{tabular}
}
\end{table}

%% file: Assets/Tables/cache_performance.tex
\begin{table}[]
\footnotesize
\centering
\renewcommand{\arraystretch}{0.9}
\caption{Average F1 score of model w/o cache and performance change under different cache strategies. Avg. denotes the average performance change over the three views.}
\label{cacheacc}

\resizebox{\linewidth}{!}{%
\begin{tabular}{llcccc|cc}
\toprule
& & \multicolumn{4}{c}{\textbf{RTMPose-L}} & \multicolumn{2}{c}{\textbf{RTMO-L}} \\
\cmidrule(lr){3-6} \cmidrule(lr){7-8}
\textbf{Movement} & \textbf{View}
& w/o & DC & RTC & Both
& w/o & RTC \\
\midrule

\multirow{4}{*}{Squat}
& Front & 0.89 & +0.04 & -0.02 & -0.05 & 0.89 & -0.02 \\
& Diag & 0.62 & +0.12 & -0.03 & -0.02 & 0.65 & +0.02 \\
& Side & 0.71 &  0.00 &  0.00 & -0.01 & 0.71 & -0.02 \\
& Avg. & -- & \textbf{+0.05} & -0.02 & -0.03 & -- & -0.01 \\
\midrule

\multirow{4}{*}{Deadlift}
& Front & 0.55 & -0.02 &  0.00 & -0.02 & 0.57 &  0.00 \\
& Diag & 0.55 & -0.03 & +0.16 & +0.03 & 0.67 &  0.00 \\
& Side & 0.89 & -0.07 & -0.09 & -0.09 & 0.82 & +0.01 \\
& Avg. & -- & -0.04 & \textbf{+0.02} & -0.03 & -- &  0.00 \\
\midrule

\multirow{4}{*}{Double Under}
& Front & 0.45 & -0.03 & -0.04 & -0.04 & 0.47 & -0.19 \\
& Diag & 0.62 & -0.03 & -0.01 & -0.04 & 0.58 & -0.09 \\
& Side & 0.80 & -0.01 & +0.01 & -0.03 & 0.79 & -0.10 \\
& Avg. & -- & -0.02 & \textbf{-0.01} & -0.04 & -- & -0.13 \\

\bottomrule
\end{tabular}
}
\end{table}

%% file: Sections/conclusion.tex
\section{Conclusion}
This paper presents \repjudge, the first knowledge-driven framework for automated rep-level judging in functional fitness, designed for edge devices. \repjudge uses LLM-assisted rule structuring to convert unstructured rulebooks into structured, pose-aligned representations using RAG and CoT reasoning. These structured representations are then used by a deterministic judge system on video-derived kinematics to evaluate rep validity and detect rep boundaries. Experimental results demonstrate effectiveness across both rule structuring and rep-level judging, with strong agreement between LLM-based evaluation and human-in-the-loop verification. The judging system achieves reliable rep-level validation under real-time constraints, and its performance is evaluated on the CFRep dataset. In particular, a dual strategy caching mechanism achieves faster-than-real-time performance (RTF $<$ 1) with reliable rep-level accuracy, delivering up to 3.36$\times$ and 15.91$\times$ speedups on resource-constrained edge devices for pre-recorded and live streaming scenarios, respectively. While the current evaluation is limited to CFRep dataset, future work will expand both movement diversity and dataset scale. Together, these results show that grounding movement evaluation in explicit, rulebook-derived representations rather than learned scoring models enables interpretable, consistent, and scalable rep-level analysis. \repjudge is therefore well-suited to complement human judging by providing repeatable evaluations beyond the real-time limits of manual observation, while remaining adaptable to evolving movement standards in both competitive and training-oriented fitness settings. 